\documentclass[runningheads]{llncs}
\usepackage{graphicx}

\usepackage{tikz}
\usepackage{comment}
\usepackage{amsmath,amssymb} 
\usepackage{color}
\usepackage{url}
\usepackage[colorlinks,
            linkcolor=red,       
            anchorcolor=blue,  
            citecolor=blue,        
            ]{hyperref}

\usepackage[accsupp]{axessibility}  

\usepackage{multirow}

\definecolor{bittersweet}{rgb}{1.0, 0.44, 0.37}
\definecolor{mygreen}{rgb}{0.29, 0.7, 0.48}

\usepackage{wrapfig}

\begin{document}
\pagestyle{headings}
\mainmatter
\def\ECCVSubNumber{1214}  

\title{GEB+: A Benchmark for Generic Event Boundary Captioning, Grounding and Retrieval} 

\titlerunning{Generic Event Boundary +}
%
\author{Yuxuan Wang\inst{1} \and
Difei Gao\inst{1} \and
Licheng Yu\inst{2} \and
Weixian Lei\inst{1} \and
Matt Feiszli\inst{2} \and
Mike Zheng Shou\inst{1}}
\authorrunning{Y. Wang et al.}
%
\institute{Show Lab, National University of Singapore\\
\and
Meta AI\\
}
\maketitle

\begin{abstract}
Cognitive science has shown that humans perceive videos in terms of events separated by the state changes of dominant subjects. State changes trigger new events and are one of the most useful among the large amount of redundant information perceived. However, previous research focuses on the overall understanding of segments without evaluating the fine-grained status changes inside.
In this paper, we introduce a new dataset called \textbf{Kinetic-GEB+}. The dataset consists of over 170k boundaries associated with captions describing status changes in the generic events in 12K videos. Upon this new dataset, we propose three tasks supporting the development of a more fine-grained, robust, and human-like understanding of videos through status changes. 
We evaluate many representative baselines in our dataset, where we also design a new \textbf{TPD (Temporal-based Pairwise Difference) Modeling} method for visual difference and achieve significant performance improvements. Besides, the results show there are still formidable challenges for current methods in the utilization of different granularities, representation of visual difference, and the accurate localization of status changes. Further analysis shows that our dataset can drive developing more powerful methods to understand status changes and thus improve video level comprehension. The dataset including \textbf{both videos and boundaries} is at
\href{https://yuxuan-w.github.io/GEB-plus/}{https://yuxuan-w.github.io/GEB-plus/}.

\end{abstract}


\begin{figure}[t]
\centering
\includegraphics[width=.9\linewidth]{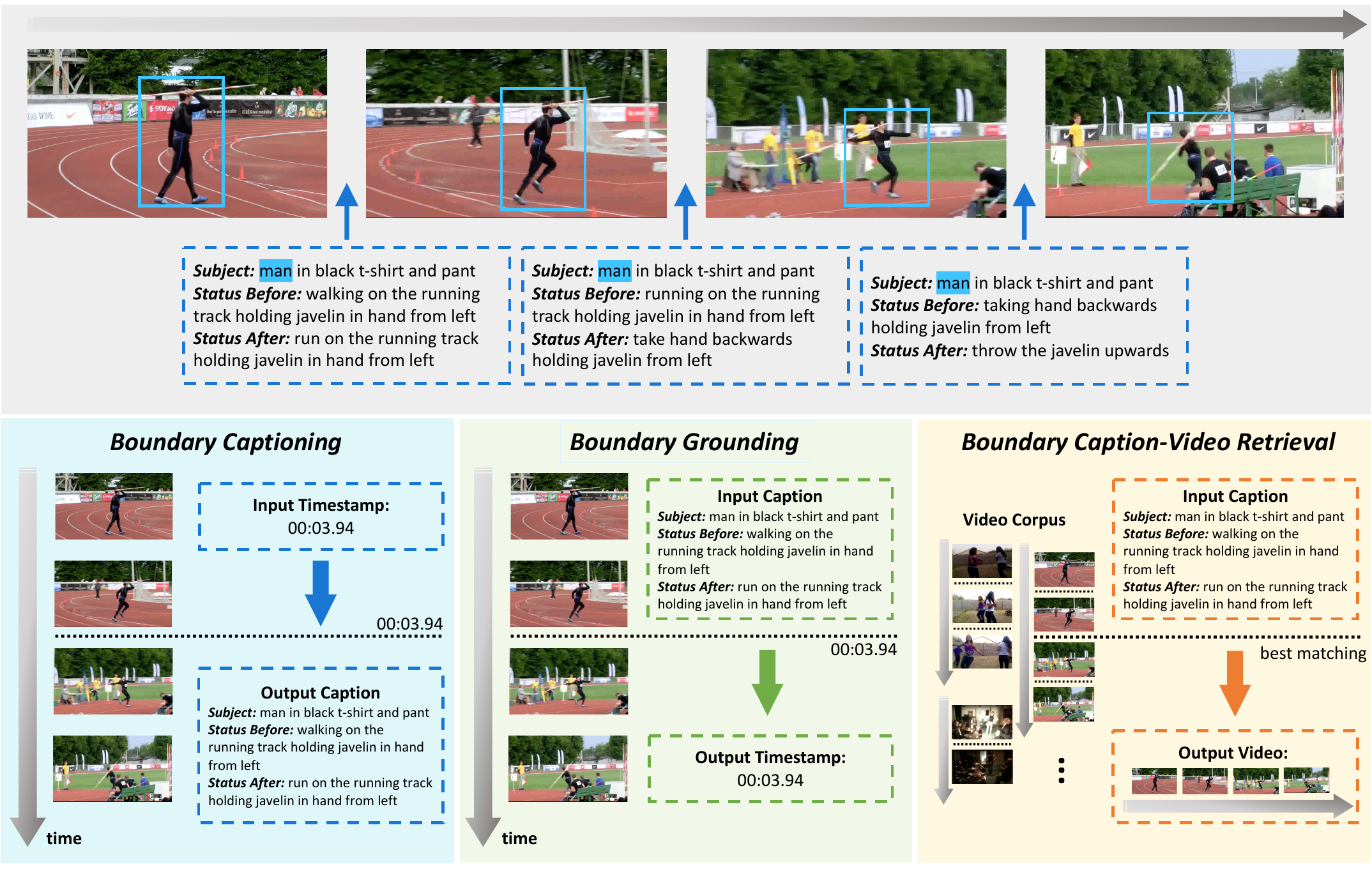}
\setlength{\abovecaptionskip}{0.5cm}
\caption{An example of generic event boundaries with captions in Kinetic-GEB+, as well as three downstream tasks designed upon the boundaries.}
\label{fig:overview}
\end{figure}

\section{Introduction}

According to cognitive science~\cite{radvansky2011event}, humans perceive videos in terms of different events, which are separated by the status changes of dominant subjects in the video. For example, in Fig.~\ref{fig:overview}, humans perceive the process of “javelin sport” by the action events such as “walking”, “running” and “throwing”. These events are triggered by the athlete’s status changes, like the instantaneous change from “walking” to “running”. The moment that instantly triggers status changes of persons, objects, or scenes often conveys useful and interesting information among a large amount of repeated, static, or regular events. Therefore, developing the understanding of the salient, instantaneous status changes is another step towards a more fine-grained and robust video understanding.
Previous works, like Dense Video Captioning~\cite{krishna2017dense,zhou2018towards,wang2021end,li2018jointly,iashin2020multi} and Video Grounding~\cite{regneri2013grounding,gao2017tall,cao2021pursuit,ge2019mac,mun2020local,yuan2019find,zeng2020dense} attempt to develop the understanding of events in video or video segments. However, these works only focus on developing an overall understanding of events rather than delving into the fine-grained status changes in the video. Other researches focusing on image level changes~\cite{park2019robust,jhamtani2018learning} employ the visual difference modeling to capture the status changes in image pairs. However, 
since the image contains only static information, the state changes exhibited by the two images involve only a few simple patterns, e.g., appear, move. These tasks are hard to evaluate the ability on understanding generic status changes.

More recently, \textit{Shou et al.}~\cite{shou2021generic} proposes Kinetic-GEBD dataset with annotated boundary timestamps for detection in Kinetic-400 videos~\cite{carreira2017quo}, where a boundary is defined as the splitter between two status of the subject. 
Though the videos in Kinetic-400~\cite{carreira2017quo} are categorized, the events selected inside are generic and mostly independent from the whole video's category.
However, in addition to letting the model predict where is the boundary, it is more important to understand why this is the boundary, which associates the visual information of boundaries with natural human languages.

Motivated by this idea, we build a new dataset called \textbf{Kinetic-GEB+ (Generic Event Boundary Captioning, Grounding and Retrieval)} which includes the video boundaries indicating status changes happening in generic events. 
For every boundary, our Kinetic-GEB+ provides the temporal location and a natural language description, which consists of the dominant \textit{Subject}, \textit{Status Before} and \textit{Status After} the boundary. 
In total, our dataset includes 176,681 boundaries in 12,434 videos selected from all categories in Kinetic-400~\cite{carreira2017quo}. 
The detailed definition of our boundary is described in Sec.~\ref{benchmark_bdycollect}. 
For future applications like AI assistant robots, with the comprehension developed from the visual status changes and natural language captions, they could understand the real time, instantaneous occurrences without hints to assist the users. 

In order to comprehensively evaluate the machine’s understanding of our boundaries, we further propose three downstream tasks shown in Fig.~\ref{fig:overview}: \textit{(1) Boundary Captioning}. Provided with the timestamp of a boundary, the machine is required to generate sentences describing the status change at the boundary. \textit{(2) Boundary Grounding}. Provided with a description of a boundary, the machine is required to locate that boundary in the video. \textit{(3) Boundary Caption-Video Retrieval}. Provided with the description of a boundary, the machine is required to retrieve the video containing that boundary from video corpus. 

In the experiment, we compare several state-of-the-art methods~\cite{lei2018tvqa,luo2020univl,zhu2020actbert,2DTAN_2020_AAAI,bain2021frozen} along with many variants on our datasets to analyze the limitation of current methods and show the challenges of the proposed tasks. 
Due to the need of visual difference for understanding the status changes, we further propose a \textbf{Temporal-based Pairwise Difference (TPD) Modeling} method representing a fine-grained visual difference before and after the boundary. This method brings a significant performance improvement. On the other hand, the results show that there are still formidable challenges for current SOTA methods in developing the comprehension of status changes.



\begin{table}[t]
\begin{center}
\caption{Comparison with most relevant Video Captioning datasets. Our \textit{Kinetic-GEB+} has comparable scale and is the only one targeting the generic boundaries, while conventional datasets focus on entire videos or video segments}
\label{table:dataset_comparison}
\resizebox{1\linewidth}{!}{
\begin{tabular}{c|rlclcrclclc}
                       & \multicolumn{1}{c}{\#Videos} &  & Video Domain  &  & \#Captions & \multicolumn{1}{c}{} & Caption Target &  & Target Type   &  & Annotation in Segments          \\ \hline
MSR-VTT                & 7,180                         &  & 20 categories &  & 200K       & \multicolumn{1}{l}{} & video          &  & generic event &  & caption                      \\
VATEX                  & 41,250                        &  & in-the-wild   &  & 825K       & \multicolumn{1}{l}{} & video          &  & action        &  & caption                      \\
Charades               & 67,000                        &  & household     &  & 20K        &                      & segment        &  & action        &  & time range + caption         \\
MSVD                   & 2,089                         &  & in-the-wild   &  & 85K        & \multicolumn{1}{l}{} & segment        &  & generic event &  & time range + caption         \\
YouCook2               & 2,000                         &  & kitchen       &  & 15K        &                      & segment        &  & action        &  & time range + caption         \\
ActivityNet Captions   & 20,000                        &  & in-the-wild   &  & 100K       &                      & segment        &  & action        &  & time range + caption         \\
\textbf{Kinetics-GEB+} & 12,434                        &  & in-the-wild   &  & 177K       &                      & boundary       &  & generic event &  & timestamp/range + caption
\end{tabular}

}
\end{center}
\end{table}
\section{Related Work}

\textbf{Video Captioning} is a conventional task with many benchmarks \cite{xu2016msr,chen-dolan-2011-collecting,wang2019vatex,krishna2017dense,zhou2018towards} established which aim to caption trimmed videos with natural language descriptions. More recently, several works~\cite{krishna2017dense,zhou2018towards,wang2021end,li2018jointly,iashin2020multi}, e.g., Dense Video Captioning~\cite{krishna2017dense}, focus on captioning the self-proposed event segments in videos. All tasks above are evaluating the overall understanding of an event, whether the event is presented in the form of a trimmed video or a video segment. In contrast, our Boundary Captioning task is to develop the comprehension of instantaneous status changes happening at boundaries, i.e., 
describing the important moment that caused a dramatic change in the state of persons, objects or scenes. As a result, there is a more urgent need for models to understand the changes in various granularity of visual concepts, e.g., action, attributes, scene status, etc. In Tab.~\ref{table:dataset_comparison}, we compare the most relevant video captioning datasets with ours.



\textbf{Image Change Captioning} is a task evaluating the ability on capturing and describing the difference between two images. There are many existing benchmarks targeting at this task. Early works~\cite{tian2013building,liu2017change} focus on changes in aerial imagery for monitoring disaster. Some other datasets~\cite{alcantarilla2018street,jhamtani2018learning} are about captioning the changes in street scenes, e.g., Spot-the-diff~\cite{jhamtani2018learning}. Recently, ~\cite{park2019robust} proposes a more challenging change caption dataset, CLEVR-change, which utilizes the CLEVR engine to construct complicated synthetic scenes to evaluate models on finding more subtle change.
One crucial limitation of previous works is that images can only present static information, thus status changes presented by two images can only involve a limited number of patterns, e.g., "appear", "disappear", "add" and "move". Towards a generic understanding of change, we extend the setting from images to videos which supports a open set of change pattern, including human action change, scene state change, etc. 


\textbf{Video Retrieval and Grounding} are both language-to-vision tasks. Given a text description of a video or event, Video Retrieval requires models to select the target video from the corpus~\cite{bain2021frozen,cheng2021improving,portillo2021straightforward}, and Video Grounding requires models to locate the target event segment (i.e. start and end boundaries) from an untrimmed video~\cite{cao2021pursuit,ge2019mac,mun2020local,yuan2019find,zeng2020dense}. 
These tasks are based on the event level understanding to find the best matching video or time span.
Compared with previous works, our Boundary Caption-Video Retrieval and Boundary Grounding tasks requires locating the two states of the subject, while traditional grounding only localizes one event. Besides, our captions are more fine-grained (describing detailed status changes) than those in traditional tasks (describing a general event).



\textbf{Generic Understanding} is a popular topic aiming to drive models from understanding predefined classes to open world vocabulary. 
Many pioneer works~\cite{bendale2015towards} propose open-set recognition tasks, which extend image classification tasks to generic understanding versions.
Some works~\cite{krishna2017dense,zhou2018towards} introduce datasets for the generic event understanding requiring models to describe videos with natural language. 
More recently, a new dataset called Kinetic-GEBD (Generic Event Boundary Detection)~\cite{shou2021generic} is proposed, which focuses on detecting the status changes between generic events. 
Our work is an extension to Kinetic-GEBD. 
We also study the boundary between events. 
However, we believe a sophisticated model should not only know where is the boundary but also understand why it is a boundary. 
Thus, this paper constructs a dataset with a large scale of boundary captions and introduces new boundary language-related tasks. 

\begin{figure}[t]
\centering
\includegraphics[width=.9\linewidth]{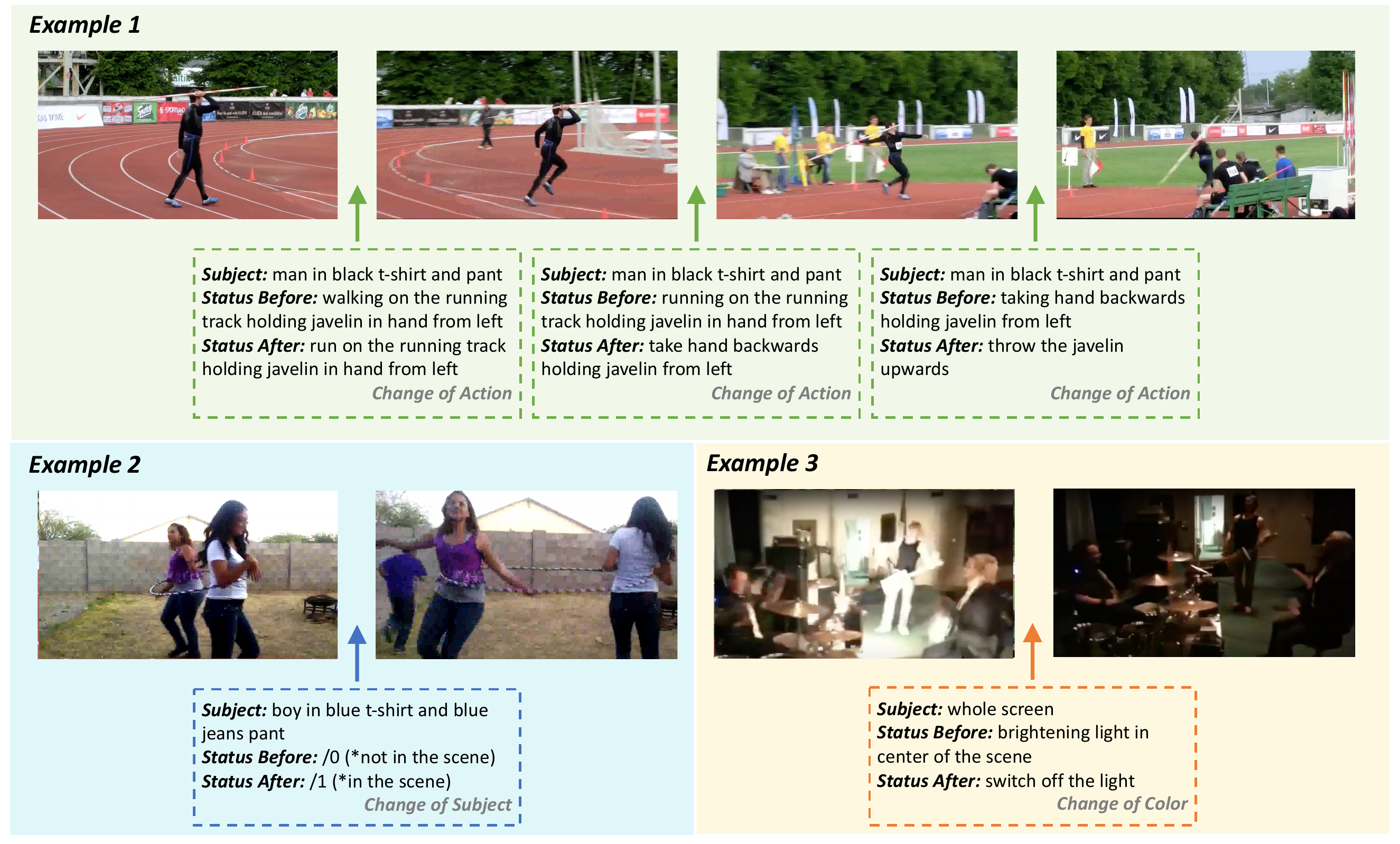}
\setlength{\abovecaptionskip}{0.5cm}
\caption{Three samples from Kinetic-GEB+. Each boundary consists of a temporal position and an associated caption, with the boundary type noted at the bottom}
\label{fig:example}
\end{figure}

\section{Benchmark Construction: Kinetic-GEB+}

To build the Kinetic-GEB+ dataset, we select 12,434 videos from the Kinetic-400 dataset~\cite{kay2017kinetics} and annotate 176,681 boundaries following a designed guideline and format. In total, our selected videos cover all the 400 categories of the Kinetic-400. It is then split into 70\% train, 15\% val and 15\% test non-overlapping sets. Several samples of boundaries are shown in Fig.~\ref{fig:example}.

\subsection{Boundary Collection}
\label{benchmark_bdycollect}

When annotating Kinetic-GEB+, one simple way would be directly captioning the boundaries in Kinetic-GEBD~\cite{shou2021generic}. 
However, our annotators did their jobs quicker when being asked to re-annotate the boundary positions than to interpret GEBD's boundaries. 
Yet, the boundaries from GEBD and GEB+ are highly consistent: when following Supp. Sec.~\ref{annotation_details:f1_consistency}
, nearly 90\%/70\% boundary positions in GEB+ reaching f1 scores higher than 0.5/0.7 with the boundaries in GEBD.

\textbf{Format and Guideline.}
Following GEBD~\cite{shou2021generic}, a boundary is defined as the splitter between two status of the subject in the video. 
Generally, we categorized our boundaries into five types: Change of Action, Change of Subject, Change of Object, Change of Color and Multiple Changes.
When annotating, we accept both single timestamps and time ranges as in~\cite{shou2021generic}, and each video is allocated to at least five annotators. 
Each annotator could independently decide whether to accept or reject the video following the criteria. 
The statistical results of annotation numbers and formats is shown in Tab.~\ref{table:annotation_num} and Tab.~\ref{table:timestamp_range}. Following~\cite{shou2021generic}, we set a minimum threshold for both temporal and spatial details’ level to ensure the consistency among different annotators. Further details are shown in Supp.


\input{tables/annotation_stat}

\subsection{Caption Collection}
In our Kinetic-GEB+, annotators are supposed to add a language description to each boundary they annotated in Sec.~\ref{benchmark_bdycollect}. To clearly and comprehensively represent humans’ understanding of the status changes, we randomly sampled 300 videos for pilot annotation to design the formats and guidelines of captioning.

\begin{figure}[t]
\centering
\includegraphics[width=.85\linewidth]{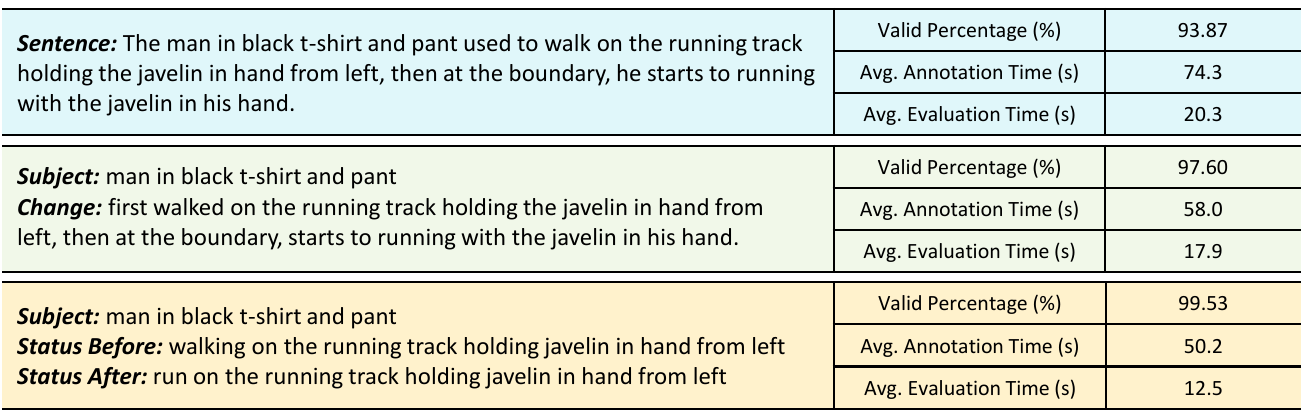}
\setlength{\abovecaptionskip}{0.5cm}
\caption{Three candidate formats of Boundary Captions and their evaluation results, respectively \textit{One-Sentence format}, \textit{Two-Item format} and \textit{Our Finalized format}}
\label{fig:format}
\end{figure}

\textbf{Format.}
Our final format of caption consists of three compulsory items: \textit{(1)} Dominant \textit{Subject} that performs the status changes. \textit{(2)} Subject's \textit{Status Before} the boundary. \textit{(3)} Subject's \textit{Status After} the boundary. In the pilot stage, we compare different versions of annotation formats as shown in Fig.~\ref{fig:format}:

\textit{One-Sentence format:} Annotators use a single sentence to describe the status change happening at each boundary. In order to obtain an open-vocab description close to daily language, we do not restrict or request anything to the expression and annotators have full autonomy in narrating. Though this format enables fluent and natural descriptions, there are significant problems in the annotations: 
\textit{(1) Ambiguity of subject}: Annotators tend to describe the subject shortly without further restriction, causing ambiguity, e.g., in a scene full of people, a short description like “a man” might indicate multiple persons.
\textit{(2) Dual changes}: Without restriction, annotators could wrongly combine two state changes of different subjects together in caption, like “Musician stops playing and an auditor starts clapping”. 
\textit{(3) Low efficiency}: Long sentences costs annotators more time to construct and takes our raters more time to understand.

\textit{Two-Items format:} To address the problems in the one-sentence format, we separate the sentence into a \textit{Subject} item and its \textit{Change} item as shown in Fig.~\ref{fig:format}. For \textit{Subject}, annotators should fill in a noun phrase. We notice that this separation makes it easier for annotators to check the singularity and specification of \textit{Subject}. Although we see that the efficiency of both annotation and evaluation are improved, this scheme still have some shortcomings:
\textit{(1) Incomplete status}: Annotators sometimes forgot to describe the status before the boundary. For example, when describing an athlete’s changing from walking to running, an annotator only filled “starts to run on the track” in \textit{Change} and forgot to mention the “walking” status before the boundary.
\textit{(2) Low efficiency}: Even though this separation improves the efficiency, the \textit{Change} item could still be too long for auditors to evaluate.
Therefore, we further separate \textit{Change} into \textit{Status Before} and \textit{Status After} to ensure the completeness of the status change description. Finally, we found this fully separated format the most efficient and robust for annotation, as shown in Fig.~\ref{fig:format}.

\begin{figure}[t]
\centering
\includegraphics[width=.9\linewidth]{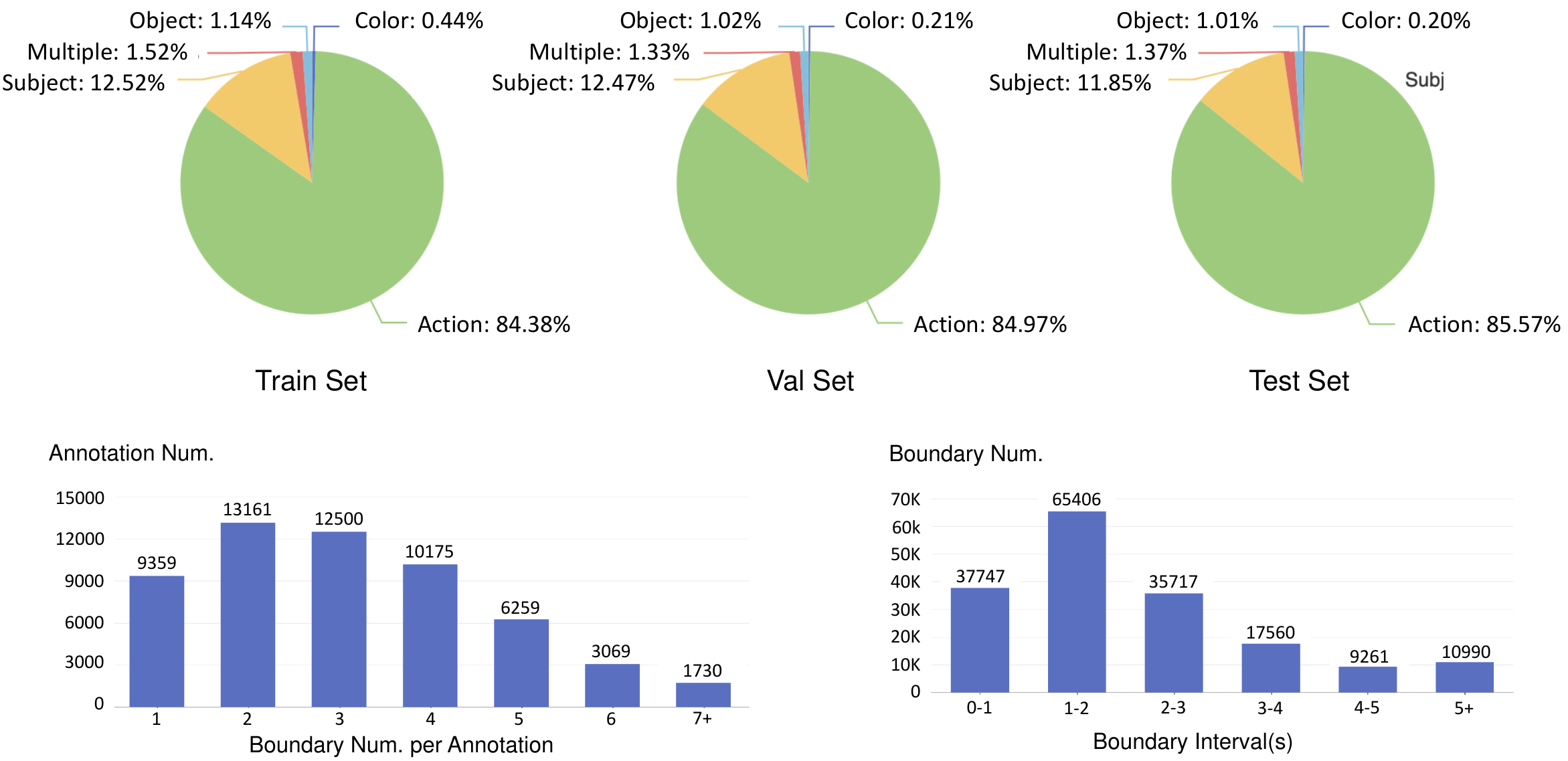}
\setlength{\abovecaptionskip}{0.9cm}
\caption{\textit{Top.} Distribution of boundary types in Train/Val/Test split. \textit{Bottom Left.} Annotation numbers versus the numbers of boundary in the annotation. \textit{Bottom Right.} Boundary numbers versus the duration of the interval before the boundary}
\label{fig:split}
\end{figure}

\textbf{Guideline.}
In our Kinetic-GEB+, the caption is defined as the reason why the annotator separates the preceding and succeeding segment of the boundary. Following the format of annotation, we brought up some specific guidance for annotating the items. Specifically, when annotating the \textit{Subject} item, annotators are required to provide distinguishable attributes of the dominant subject. However, in complex cases where the subject is difficult to describe without ambiguity  (e.g. many people dressing similarly in the scene), the annotator could just describe some attributes to avoid verbose descriptions.

When annotating \textit{Status Before} and \textit{Status After}, annotators are required to limit their attention to the time range between the proceeding boundary and succeeding boundary, thus to ensure all the status changes in the same video are at the same temporal level. To further improve the consistency of expressions, we employ the symbol \textit{/1} and \textit{/0} to represent the appearance and disappearance of a subject in the scene, as shown in \textit{Example 2} of Fig.~\ref{fig:example}. Finally, we embrace all the tenses only if the annotators feel natural. In this way, we ensure the specification of descriptions while keeping their naturalness.

\begin{figure}[t]
\centering
\includegraphics[width=.99\linewidth]{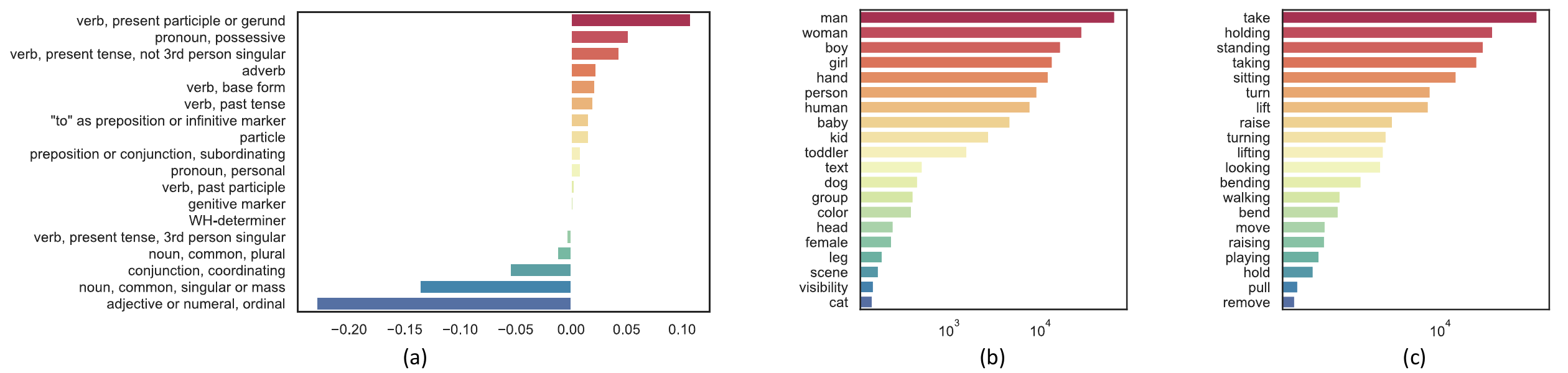}
\setlength{\abovecaptionskip}{0.5cm}
\caption{\textit{(a)} The parts of speech distribution of the \textit{Status Before} and \textit{Status After} compared with that of the Subject part. The two status parts contains more verbs and focus more on motions. \textit{(b)} The 20 most frequent nouns in \textit{Subject}. \textit{(c)} The 20 most frequent verbs in \textit{Status Before} and \textit{Status After}}
\label{fig:caption_stat}
\end{figure}




\subsection{Statistics}

\textbf{Splitting.}
When splitting our Kinetic-GEB+ into train, validation and test sets, the boundary type is the most important characteristic of consistency, since it determines which granularities the model should depend on to understand the state change. We allocate videos containing different types of boundaries by proportion to ensure the distribution is the same in all splits. The final distribution is shown in Fig.~\ref{fig:split}, where we see the distribution is consistent in three splits. More details of splitting criteria is discussed in Supp. Sec.~\ref{supp-sec:anno}.

\textbf{Boundary number.}
To quantify the density of annotated boundaries, we make a statistics of the boundary number in each piece of annotation. Notably, due to the variant understanding of annotator, annotations of the same video could have different numbers of boundaries. The bottom left side of Fig.~\ref{fig:split} shows the counts of annotations versus their boundary numbers, from which we could see that most of annotations have 1 to 4 boundaries. 


\textbf{Boundary interval.}
Furthermore, to investigate the duration of events located between two boundaries, we conduct the statistics on the length of intervals. For the first boundary in the video, we take the distance to the start of the video as its interval duration. The result is shown in the bottom right side of Fig.~\ref{fig:split} which is similar to the statistic of boundary numbers.

\textbf{Part of speech comparison in caption.}
For the captions in our dataset, we first analyze and compare the part of speech distributions in the subject and two status parts. In Fig.~\ref{fig:caption_stat}(a), the comparison result indicates that the status parts contain more verbs and focus more on actions than the subject part. On the other hand, the subject part includes more nouns and adjectives than the two status parts, suggesting it focuses more on appearance information.

\textbf{Frequent subjects and actions in caption.}
To further analyze the different aspects of information in the three parts. In Fig.~\ref{fig:caption_stat}(b)(c), we extract the first noun in every \textit{Subject} as well as the first verb in all \textit{Status Before} and \textit{Status After}, and then illustrate the 20 most frequent words. Same with Kinetic-400, we see that both the nouns and verbs in our datasets are mainly correlated with the appearance and motions of humans. This conforms to the scenarios of practical application, since humans are also the dominant subject in most of the scenes.

\subsection{Adjustment for downstream tasks}
\label{benchmark_adjust}

For downstream tasks, we select one annotator whose labeled boundaries are mostly consistent with others to reduce noise and duplication. Then, we use these boundaries’ timestamps as the anchors to merge other annotators’ captions, preserving the diversity of different opinions. Thus, one video corresponds to multiple boundaries, and each boundary could be with multiple captions. Finally, this selection includes 40k anchors from all videos. 
Furthermore, we find two different boundaries in the same video could be occasionally too similar in semantics for even humans to tell. For \textit{Boundary Grounding}, we mark these pair of boundaries as equal in the ground truth. More details are discussed in Supp.



\section{Experiments}
Kinetic-GEB+ dataset enables us to benchmark how well current mainstream methods could comprehend the instantaneous status changes in videos. For each task, we implement and compare among SOTA models with our modifications, as well as further explorations on ablation and visual difference modeling methods. 

\begin{figure}[t]
\centering
\includegraphics[width=0.9\linewidth]{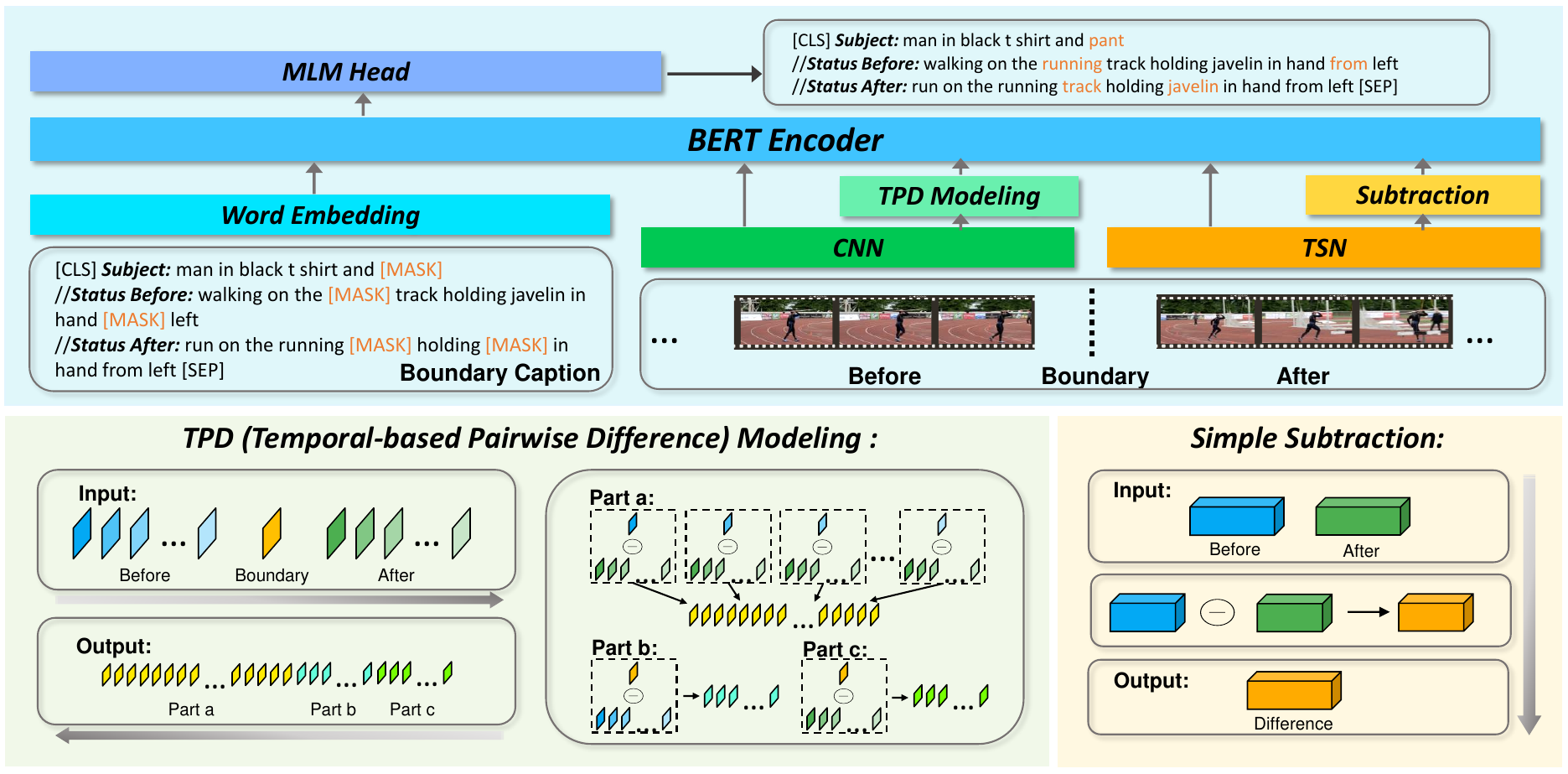}
\caption{\textit{Top}. A general modification for BERT model showing on \textbf{ActBERT-revised}. \textit{Bottom}. Our difference modeling methods designed for BERT model}
\label{fig:diff_modeling}
\end{figure}

\subsection{Methods}
\label{experiment_methods}
\textbf{Granularities of Input Features.}
We extract multiple granularities of features and utilize different combinations of them in experiments. Given each boundary, we sampled multiple frames before and after the timestamp and one frame at the timestamp for further extraction.

Our features include: \textit{(1) ResNet}: Firstly, we extract a 1024 dimensional ResNet-roi feature using ResNet~\cite{he2016deep} followed by Region of Interest (RoI) pooling. Then we extract another ResNet-conv feature to fit~\cite{park2019robust}: We sample one frame before and another frame after the boundary, then extract the Conv features from the two frames. \textit{(2) TSN}: For frames before and after the timestamp, we extract a 2048 dimensional TSN feature for the before and after snippets using pre-trained TSN~\cite{wang2018temporal} network. \textit{(3) Faster R-CNN}: For every sampled frame, we employ Faster R-CNN~\cite{ren2015faster} to extract the 1024 dimensional R-CNN feature by selecting 20 objects with highest confidence. \textit{(4) C3D}: Similar to the TSN feature, we extract 4096 dimensional C3D features with pre-trained C3D~\cite{tran2015learning} network for the before and after snippets to fit~\cite{2DTAN_2020_AAAI}.

These features are categorized into two granularities: \textit{Instant-granularity} features extracted from the instantaneous appearance in a single frame, such as the R-CNN and ResNet features, are to provide fine-grained visual information of instants. \textit{Event-granularity} features, like the TSN and C3D feature, could provide an overall representation of appearance and motion information in event snippets. We assume that developing a fine-grained understanding of status changes requires both the granularities.

\textbf{Backbones.}
We implement the following backbones with various adoption and modification according to the tasks: 
\textit{(1) CNN+LSTM}: A rudimentary backbone that simply uses a vanilla LSTM which takes the CNNs extracted features as input. The output of LSTM is mapped to caption tokens in Boundary Captioning, or is max-pooled to be the matching score in other two tasks. 
\textit{(2) Dual Dynamic Attention Model (DUDA)}: The baseline method in~\cite{park2019robust} which consists of a CNN-based Change Detector and a LSTM-based Dynamic Speaker. Besides, it utilizes a simple visual difference modeling by subtraction. 
\textit{(3) ActionBERT-revised}: A one-stream BERT architecture using early fusion from~\cite{zhu2020actbert}. We modify the structure by applying difference modeling after the embedding and employing different feature combinations. 
\textit{(4) UniVL-revised}: A two-stream BERT architecture from~\cite{luo2020univl}, which includes a caption encoder, a context encoder and a cross-encoder for late fusion. We apply difference modeling to the context encoder with different feature combinations. 
\textit{(5) FROZEN-revised}: A two-stream BERT architecture from~\cite{bain2021frozen}, which includes a caption encoder and a context encoder with no fusion. The revision is the same as \textit{UniVL-revised}. 
\textit{(6) TVQA}: The baseline method in~\cite{lei2018tvqa}, where we remove all the “answer” substreams and process each visual granularity with one stream.
\textit{(7) 2D-TAN}: The baseline method in~\cite{2DTAN_2020_AAAI}, where we only keep the diagonal elements in the 2D map.

\textbf{Visual Difference Modeling.}
Developing a fine-grained understanding of status changes at the boundary requires visual difference information. 
Most existing methods are focused on image-pair differences~\cite{park2019robust}, where the difference is obtained by simply subtracting the “before” image from the “after” image. 
A simple inference of this method on video tasks is by pooling the sampled frames then doing subtraction. 
However, this method only provides an event-granularity representation of the visual difference between the before and after snippets, and thus loses the instant-granularity visual differences. 

To address this problem, we design a new method of \textbf{Temporal-based Pairwise Difference (TPD) Modeling} for BERT models. 
As shown in Fig.~\ref{fig:diff_modeling}, we first compute the pairwise subtraction between the embedding of frames in "before" and "after" as \textit{Part a}, where the embeddings of the frames are sampled in Sec.~\ref{experiment_methods}.
This provides us a fine-grained and wide-viewing visual comparison between the status before and after. 
To represent the visual difference between the boundary and other sampled timestamps, we further compute \textit{Part b} and \textit{Part c}, which includes the pairwise subtraction between the frame embeddings at the boundary and that before or after the boundary. 
Finally, we concatenate all these differences together as the output of \textit{TPD Modeling}.

The advantage of our \textit{TPD Modeling} is that, compared with previous methods designed for image tasks, it provides multiple granularities of information and ensures the fine-grained representation of visual differences. 
In the ablation study of Boundary Captioning, we design an experiment to explore the difference modeling methods and verify our perceptions.

\begin{table}[t]
\begin{center}
\caption{Performance of Different Methods in Boundary Captioning. For \textit{UniVL-revised} and \textit{ActBERT-revised}, we apply the \textit{TPD Modeling} and take the "ResNet-roi+TSN" combination as input feature}
\label{table:captioning_sota}
\resizebox{1\linewidth}{!}{
\begin{tabular}{llll|lcccllcccllccc}
\hline
\multicolumn{4}{c|}{\multirow{2}{*}{Method}}      & \multicolumn{4}{c}{CIDEr}                                          &  & \multicolumn{4}{c}{SPICE}                                         &  & \multicolumn{4}{c}{ROUGE\_L}                                      \\
\multicolumn{4}{c|}{}                             & Avg.           & Sub.            & Bef.           & Aft.           &  & Avg.           & Sub.           & Bef.           & Aft.           &  & Avg.           & Sub.           & Bef.           & Aft.           \\ \cline{1-8} \cline{10-13} \cline{15-18} 
\multicolumn{4}{l|}{CNN+LSTM}                     & 49.73          & 80.11           & 34.39          & 34.69          &  & 13.62          & 18.84          & 9.92           & 12.10          &  & 26.46          & 39.77          & 20.77          & 18.83          \\
\multicolumn{4}{l|}{DUDA}     & 58.56          & \textbf{104.41} & 47.12          & 24.14          &  & 16.34          & \textbf{21.72} & 14.63          & 12.68          &  & 27.57          & \textbf{42.76} & 21.76          & 18.18          \\
\multicolumn{4}{l|}{UniVL-revised (two-stream)}   & 65.74          & 91.51           & 56.58          & 49.13          &  & 18.06          & 21.08          & 17.06          & 16.05          &  & 26.12          & 40.67          & 19.42          & 18.28          \\
\multicolumn{4}{l|}{ActBERT-revised (one-stream)} & \textbf{74.71} & 85.33           & \textbf{75.98} & \textbf{62.82} &  & \textbf{19.52} & 20.10          & \textbf{20.66} & \textbf{17.81} &  & \textbf{28.15} & 39.16          & \textbf{23.70} & \textbf{21.60} \\ \hline
\end{tabular}
}
\end{center}
\end{table}
\subsection{Boundary Captioning}
\label{experiment_captioning}
For Boundary Captioning, we first implement and compare the performance of \textit{CNN+LSTM}, \textit{DUDA}, \textit{UniVL-revised} and \textit{ActBERT-revised}. To further explore how different input granularities support the understanding, we design a series of ablation studies using \textit{ActBERT-revised} for all combinations of input features. In these two experiments, we apply our \textit{TPD Modeling} as shown in Fig.~\ref{fig:diff_modeling}.

To find the best schemes to represent visual difference, we further compare the performances of three schemes on \textit{ActBERT-revised}: \textit{(1)} Embedding with no difference modeling. \textit{(2)} Max-pooling the frames before and after the boundary and simply subtracting one from another, which is inferred from the current method in~\cite{park2019robust}. \textit{(3)} Using \textit{TPD Modeling} to represent the visual differences. In Supp. Sec.~\ref{supp-sec:more_experiments}, we conduct an ablation study of different parts of \textit{TPD Modeling} and explore on several other methods for visual difference representation.

\begin{table}[t]
\begin{center}
\caption{\textit{Upper}. Ablation study results of the Boundary Captioning utilizing \textit{ActBERT-revised} with \textit{TPD Modeling} employed to all rows with "ResNet-roi". \textit{Lower}. The performance comparison of visual difference modeling methods, where the \textit{TPD Modeling} is employed to the last row}
\label{table:captioning_ablation}
\resizebox{1\linewidth}{!}{
\begin{tabular}{l|cccccccccccccc}
\hline
\multicolumn{1}{c|}{\multirow{2}{*}{Input Granularity}} & \multicolumn{4}{c}{CIDEr}                                         &                      & \multicolumn{4}{c}{SPICE}                                         &                      & \multicolumn{4}{c}{ROUGE\_L}                                      \\
\multicolumn{1}{c|}{}                                   & Avg.           & Sub.           & Bef.           & Aft.           & \multicolumn{1}{l}{} & Avg.           & Sub.           & Bef.           & Aft.           & \multicolumn{1}{l}{} & Avg.           & Sub.           & Bef.           & Aft.           \\ \cline{1-5} \cline{7-10} \cline{12-15} 
ResNet-roi                                             & 51.93          & 67.79          & 46.59          & 41.42          &                      & 14.30          & 16.01          & 13.54          & 13.34          &                      & 24.20          & 35.42          & 19.04          & 18.13          \\
ResNet-conv                                                  & 66.18          & \textbf{96.86} & 54.77          & 46.91          &                      & 17.07          & 20.58          & 15.82          & 14.8           &                      & 26.30          & 40.38          & 19.71          & 18.82          \\
TSN                                                     & 70.80          & 92.54          & 65.64          & 54.21          &                      & 19.00          & \textbf{20.97} & 18.98          & 17.04          &                      & 26.89          & \textbf{40.53} & 20.82          & 19.32          \\
ResNet-roi + ResNet-conv                                    & 56.64          & 83.82          & 45.64          & 40.45          &                      & 15.68          & 19.17          & 13.77          & 14.1           &                      & 25.46          & 38.64          & 19.26          & 18.47          \\
ResNet-conv + TSN                                            & 69.58          & 83.56          & 68.88          & 56.3           &                      & 18.95          & 20.15          & 19.51          & 17.2           &                      & 27.14          & 38.52          & 22.36          & 20.53          \\
ResNet-roi + TSN                                       & \textbf{74.71} & 85.33          & \textbf{75.98} & \textbf{62.82} &                      & \textbf{19.52} & 20.10          & \textbf{20.66} & \textbf{17.81} &                      & \textbf{28.15} & 39.16          & \textbf{23.70} & \textbf{21.60} \\
ResNet-roi + ResNet-conv + TSN                              & 65.83          & 80.9           & 63.22          & 53.38          &                      & 18.69          & 19.37          & 19.25          & 17.46          &                      & 26.84          & 37.82          & 22.11          & 20.59          \\ \cline{1-5} \cline{7-10} \cline{12-15} 
ResNet-roi + TSN (w/o Diff.)                           & 67.38          & \textbf{85.59} & 63.06          & 53.49          &                      & 18.47          & 19.84          & 18.69          & 16.87          & \textbf{}            & 24.23          & 31.65          & 21.14          & 19.90          \\
ResNet-roi + TSN (simple)                              & 67.75          & 85.31          & 64.28          & 53.65          &                      & 18.96          & \textbf{20.35} & 19.13          & 17.39          &                      & 26.78          & 39.14          & 21.20          & 20.00          \\
ResNet-roi + TSN (TPD)                                      & \textbf{74.71} & 85.33          & \textbf{75.98} & \textbf{62.82} &                      & \textbf{19.52} & 20.10          & \textbf{20.66} & \textbf{17.81} &                      & \textbf{28.15} & \textbf{39.16} & \textbf{23.70} & \textbf{21.60} \\ \hline
\end{tabular}
}
\end{center}
\end{table}
\textbf{Implementation.}
For \textit{CNN+LSTM} and \textit{DUDA}, we utilize the ResNet-conv feature following~\cite{park2019robust}. For \textit{UniVL-revised} and \textit{ActBERT-revised}, we utilize the ResNet-roi feature and TSN feature described in Sec.~\ref{experiment_methods}, where the sampling range is from the preceding boundary to the succeeding boundary. 
In evaluation, we separate the prediction into the three items, and then compute the similarity score of each item with the ground truth. After that, we employ CIDEr~\cite{vedantam2015cider}, SPICE~\cite{anderson2016spice} and ROUGE\_L~\cite{lin2004rouge} as evaluation metrics, which are widely utilized in image and video captioning benchmarks. Further details are discussed in Supp.

\textbf{Result.}
From Tab.~\ref{table:captioning_sota}, we see that the \textit{ActBERT-revised} backbone performs the best. However, the results in are still far from satisfactory, thus we further analyze the challenges of our task through the result in Tab.~\ref{table:captioning_ablation}:

\textit{Accurate captioning of the status changes requires both the instant and event granularities.} First, the event-granularity features perform as the base of the understanding. In Tab.~\ref{table:captioning_ablation}, the "ResNet-roi+TSN" combination outperforms all the groups using only the instant-granularity features (e.g. the combinations of ResNet features). Second, a proper usage of the instant-granularity features could help to enrich the understanding. 
As in Tab.~\ref{table:captioning_ablation}, the "ResNet-roi+TSN" combination outperforms the single TSN feature.

\textit{Our task requires adaptive usage of different granularities.} Machines need to know when to look at which granularity. Simply assembling different features together could sometimes disturb the attention resulting in worse performance. 
In Tab.~\ref{table:captioning_ablation}, when only utilizing the TSN feature, the performance is better than using either "ResNet-roi+TSN" or "ResNet-roi+ResNet-conv+TSN" combination. 

\textit{Understanding the status changes requires effective modeling of visual differences} In the comparison of difference modeling schemes in Tab.~\ref{table:captioning_ablation}, the plain embedding without difference modeling performs the worst, while the utilization of simple-subtraction difference modeling brings little improvement to the performance. At the same time, the group with our \textit{TPD Modeling} method significantly outperforms others. This gap in performance conforms to our perspective that learning a fine-grained understanding of status changes requires not only an overall but also a fine-grained representation of visual differences.

\begin{table}[t]
\begin{center}
\caption{Performance comparison among different methods in Boundary Grounding. For \textit{UniVL-revised} and \textit{ActBERT-revised}, we apply \textit{TPD Modeling} and take the “ResNet-roi+TSN" combination as input feature}
\label{table:grounding_sota}
\resizebox{.8\linewidth}{!}{
\begin{tabular}{l|clclclclclclclclc}
\hline
\multicolumn{1}{c|}{\multirow{2}{*}{Method}} & \multicolumn{17}{c}{Threshold (s)}                                                                                                                                                                                                          \\
\multicolumn{1}{c|}{}                        & 0.1           &           & 0.2           &           & 0.5            &           & 1              &           & 1.5            &           & 2              &           & 2.5            &           & 3              &  & Avg.           \\ \cline{1-2} \cline{4-4} \cline{6-6} \cline{8-8} \cline{10-10} \cline{12-12} \cline{14-14} \cline{16-16} \cline{18-18} 
Random Guess                                 & 2.14          & \textbf{} & 4.56          & \textbf{} & 11.46          & \textbf{} & 22.81          & \textbf{} & 31.63          & \textbf{} & 40.43          & \textbf{} & 48.06          & \textbf{} & 54.37          &  & 26.93          \\
TVQA                                         & 2.60          & \textbf{} & 5.30          & \textbf{} & 12.90          & \textbf{} & 23.73          & \textbf{} & 32.94          & \textbf{} & 41.33          & \textbf{} & 48.56          & \textbf{} & 55.17          &  & 27.82          \\
2D-TAN                                       & 2.91          & \textbf{} & 6.32          & \textbf{} & 15.04          & \textbf{} & 26.95          & \textbf{} & 36.94          & \textbf{} & 45.34          & \textbf{} & 51.87          & \textbf{} & 58.22          &  & 30.45          \\
ActBERT-revised                 & 3.12          & \textbf{} & 6.14          & \textbf{} & 14.79          & \textbf{} & 26.78          & \textbf{} & 36.61          & \textbf{} & 45.45          & \textbf{} & 52.99          & \textbf{} & 59.41          &  & 30.66          \\
FROZEN-revised                  & \textbf{4.28} & \textbf{} & \textbf{8.54} & \textbf{} & 18.33          & \textbf{} & \textbf{31.04} & \textbf{} & \textbf{40.48} & \textbf{} & 47.86          & \textbf{} & 54.81          & \textbf{} & 61.45          &  & \textbf{33.35} \\
FROZEN-revised-GEBD             & 4.20          & \textbf{} & 8.48          & \textbf{} & \textbf{18.49} & \textbf{} & 29.91          & \textbf{} & 39.54          & \textbf{} & \textbf{48.37} & \textbf{} & \textbf{55.29} & \textbf{} & \textbf{61.55} &  & 33.23          \\ \hline
\end{tabular}
}
\end{center}
\end{table}
\begin{table}[t]
\begin{center}
\caption{Performance comparison of different methods in Boundary Caption-Video Retrieval. For \textit{FROZEN-revised}, we add another group without difference modeling}
\label{table:retrieval_sota}
\resizebox{0.7\linewidth}{!}{
\begin{tabular}{l|lllllllll}
\hline
\multicolumn{1}{c|}{Method}          & \multicolumn{1}{c}{mAP} &           & \multicolumn{1}{c}{R@1} &           & \multicolumn{1}{c}{R@5} &           & R@10           &           & R@50           \\ \cline{1-2} \cline{4-4} \cline{6-6} \cline{8-8} \cline{10-10} 
Random                               & 0.39                    &           & 0.05                    &           & 0.23                    &           & 0.44           &           & 2.52           \\
CNN+LSTM                             & 9.25                    &           & 4.08                    &           & 12.49                   &           & 19.53          &           & 42.26          \\
ActBERT-revised (one-stream)         & 19.14                   &           & 9.52                    &           & 28.89                   &           & 40.14          &           & 64.50          \\
FROZEN-revised (two-stream)          & \textbf{23.39}          & \textbf{} & \textbf{12.80}          & \textbf{} & \textbf{34.81}          & \textbf{} & \textbf{45.66} & \textbf{} & \textbf{68.10} \\
FROZEN-revised (two-stream) w/o diff & 22.44                   &           & 12.12                   &           & 33.42                   &           & 43.89          &           & 65.61          \\ \hline
\end{tabular}
}
\end{center}
\end{table}
\subsection{Boundary Grounding}
In Boundary Grounding, we compare the performance of four backbones: \textit{TVQA}, \textit{2D-TAN}, \textit{FROZEN-revised} and \textit{ActionBERT-revised}. Given a video and a caption query, the model computes the matching scores of each candidate sampled from the video, followed with post-processing to finalize the prediction.

\textbf{Implementation.} 
In the training period, we use the ground truth boundaries processed in Sec.~\ref{benchmark_adjust} and their timestamps. In testing, we employ two strategies to sample the timestamp candidates for groups as specified in their suffix. More details are discussed in Supp.
For \textit{2D-TAN}, we utilize the C3D feature as in~\cite{2DTAN_2020_AAAI}. For \textit{TVQA}, we utilize the R-CNN and ResNet-roi features as context. Besides, we build the triplets consisting of one positive and two negative pairs, and then compute the cross-entropy loss for each triplet in training. In \textit{ActBERT-revised} and \textit{FROZEN-revised}, we apply the contrastive loss in~\cite{bain2021frozen} as objective and implemented a batch-randomed sequential sampler in training. The batch-randomed sampler allocates the boundaries in the same video to the same batch, encouraging the model to learn the visual differences within videos.

After the models generate the matching scores of all candidate timestamps, we apply the Laplace-of-Gaussian filter in~\cite{shou2021generic} to derive local maximas of the scores. Then we select the top-$k$ maximas as final prediction, where $k$ is subject to the statistical number of ground truth timestamps marked in Sec.~\ref{benchmark_adjust}. To evaluate the accuracy of the prediction, we compute F1 scores based on the absolute distance between ground truth timestamps and predicted timestamps, with the threshold varying from 0.1s to 3s. Further details are discussed in Supp.

\textbf{Result.}
We see that \textit{FROZEN-revised} performs the best in the comparison of SOTA methods in Tab.~\ref{table:grounding_sota}. However, all the SOTA methods struggle when the threshold is less than $1$s, indicating that \textit{improving the temporal resolution of understanding is still a main challenge of our task}. Future improvements still need to focus on how to delve deeper into the temporal details and prevent the models from taking a glance and learning a rough impression of status changes.

\subsection{Boundary Caption-Video Retrieval}
We implement and compare the performance of the \textit{CNN+LSTM}, \textit{FROZEN-revised} and \textit{ActionBERT-revised} backbones. Same as in Boundary Grounding, the backbones is to compute the matching score between the query and context.

\textbf{Implementation.}
In order to find the target video from the corpus, each query is to be tried to match with every boundary candidate from all videos. Considering the corpus size, we only apply the baseline in~\cite{shou2021generic} to generate the boundary candidates. When implementing \textit{CNN+LSTM}, we take the R-CNN and ResNet-roi features as visual contexts. For \textit{FROZEN-revised} and \textit{ActBERT-revised}, we utilize the same configuration with Boundary Grounding. To evaluate the retrieval accuracy, for each query, we sort all the videos by the highest scores of their boundary candidates and then compute the mAP and recall metrics.

\textbf{Result.}
In Tab.~\ref{table:retrieval_sota}, \textit{FROZEN-revised} with difference modeling performs the best, but the performance gap is significantly smaller than in Boundary Grounding, suggesting that this video-level retrieval task relies less on the fine-grained visual differences. It is natural since the overall video-level understanding is already enough to distinguish the target among different videos. 

\section{Conclusion}
\label{conclusion}
In this paper, we have introduced our new dataset \textit{Kinetic-GEB+} with the methods of benchmark construction and proposed three tasks that aim to develop a more fine-grained, robust and human-like understanding of videos based on status changes. We further explore the challenges with designed experiments, where we design a new \textit{Temporal-based Pairwise Difference (TPD)} modeling method to represent visual differences and obtain significant improvement in performance. 
Concluding the results from the experiments, we summarize the challenges of our benchmarks as three parts: \textit{(1)} How to adaptively utilize multiple granularities of features and exclude the disturbance. \textit{(2)} How to effectively represent the visual differences around the boundary. \textit{(3)} How to improve the temporal resolution of understanding. 
We believe our work could be a stepping stone for the following works to develop more powerful methods to understand status changes and thus improve video-level comprehension. 


~\\~\\
\noindent
\textbf{Acknowledgements.} This project is supported by the National Research Foundation, Singapore under its NRFF Award NRF-NRFF13-2021-0008, and Mike Zheng Shou's Start-Up Grant from NUS. The computational work for this article was partially performed on resources of the National Supercomputing Centre, Singapore.

\clearpage
%
%

\section*{Supp.: Overview}

In the supplementary material, we provide 
more details of annotations (Sec.~\ref{supp-sec:anno}) and more implementation details of the baselines (Sec.~\ref{supp-sec:imp-detail}). Moreover, we conduct more experiments of Boundary Captioning and Grounding for more visual difference representation methods as well as the ablation study for our \textit{TPD Modeling} method (Sec.~\ref{supp-sec:more_experiments}). Finally, we release some common failure cases in our prediction of Boundary Captioning and further discussion on the benchmark (Sec.~\ref{supp-sec:more_discusstion}).

\section{More Details of Annotations}
\label{supp-sec:anno}

\subsection{Boundary Definition}

\subsubsection{Specifying the level of details.}
A great number of our video sources from \textit{Kinetic-400} contain more than one actor or object with different levels of status changes, and different annotators could have high-variance opinions on the boundary positions. According to~\cite{shou2021generic}, to reduce the variance among annotators, the highest priority is to specify the level of the spatial and temporal details we take into consideration. For the level of spatial details, we only focus on the event changes that are performed by dominant subjects. Specifically, in the \textit{Example 2} of Fig.~\ref{fig:example}, the two girls are repeating the same event, the status of which is unchanged (no event boundary). Instead, the boy performs different events before and after the marked timestamp. For the level of temporal details, we only consider the “one-level-deeper” granularity as in~\cite{shou2021generic}. By specifying this, we ensure that most of the boundaries are in the same granularity, rendering it possible for annotators to basically reach an agreement on the boundary location without predefined classes.

\subsubsection{Embracing the Ambiguity.}
Knowing the specified level of details, however, different annotators could still have some disagreements on the dominant subjects and the “one-step-deeper” granularity events. Following~\cite{shou2021generic}, we embrace this varsity when annotating. For each video, we take all the annotations as correct. Then we supervise the consistency among different annotations towards the same video by calculating the F1 score in Sec.~\ref{annotation_details:f1_consistency}.

\begin{figure}[t]
\centering
\includegraphics[width=.6\linewidth]{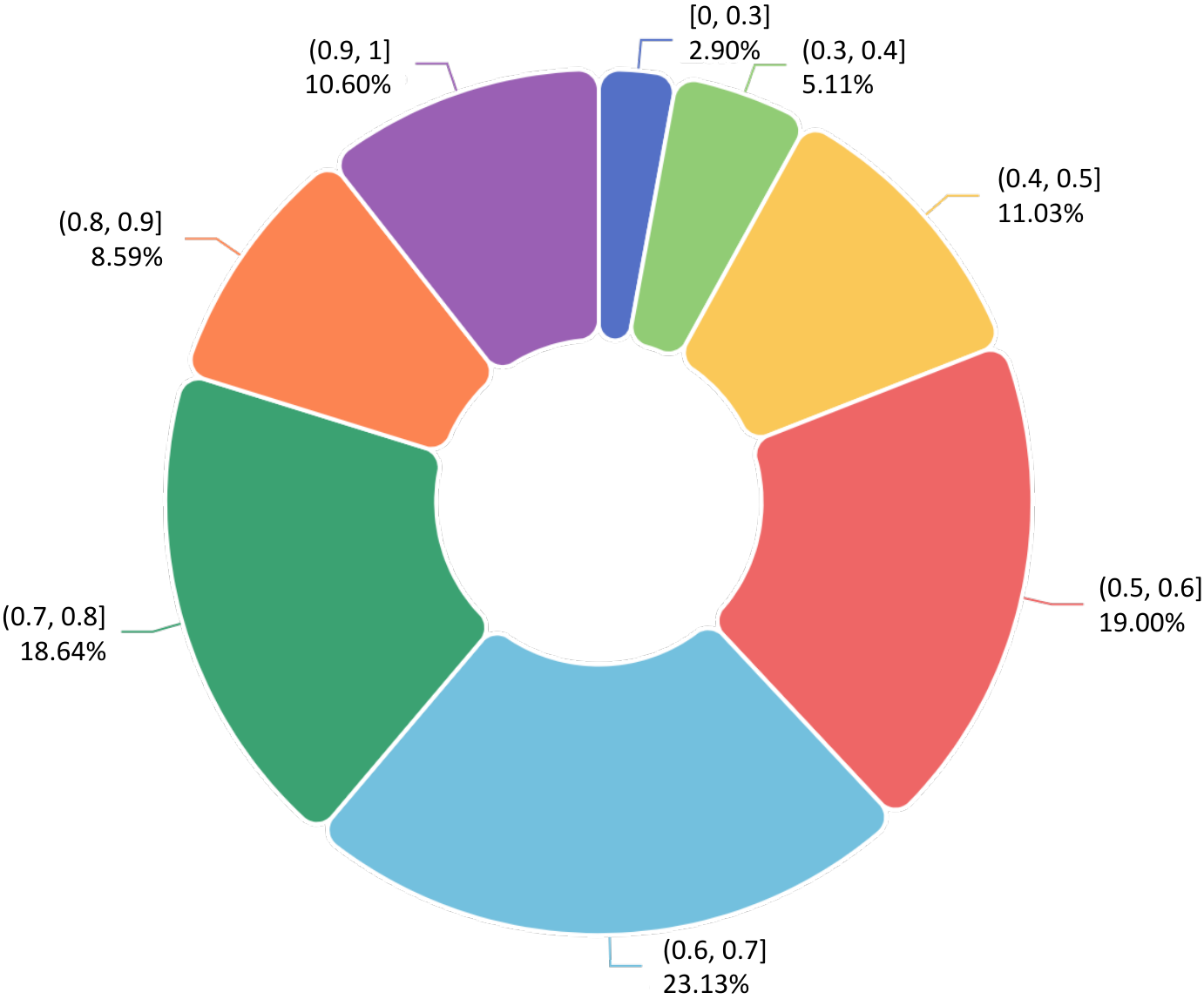}
\caption{Distribution of consistency F1 scores in all annotations. We first compute the F1 scores with different thresholds from 0.2s to 1s, and then average the scores in all thresholds as the final score}
\label{fig:consistency_f1}
\end{figure}

\begin{figure}[t]
\centering
\includegraphics[width=.99\linewidth]{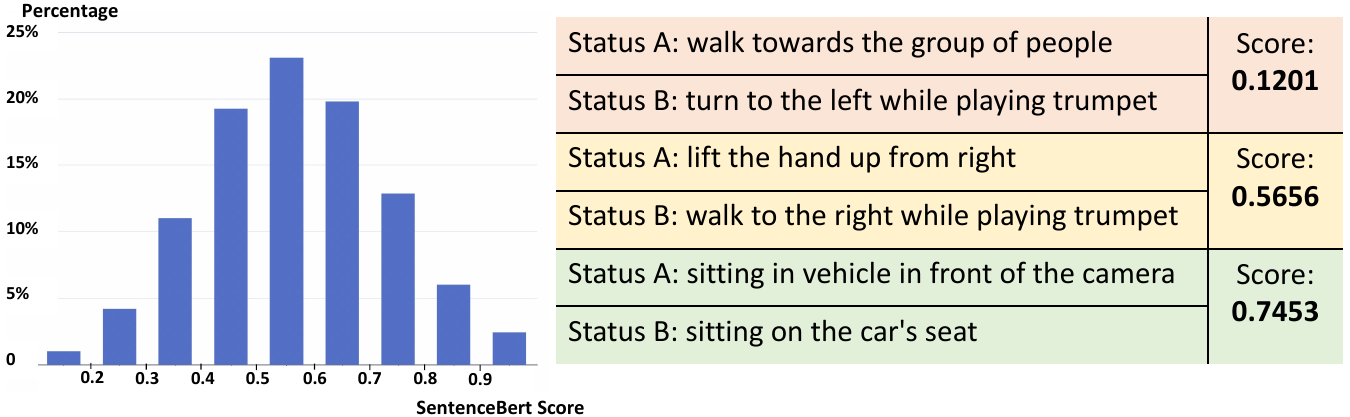}
\caption{\textit{Left.} Distribution of Sentence-BERT scores for the sampled captions in the same videos. \textit{Right.} Intuitive examples of different level of Sentence-BERT scores selected from captions in the same videos in Kinetic-GEB+.}
\label{fig:caption_similarity}
\end{figure}

\subsection{Quality Assurance}
\label{Supp-sec:quality_assurance}

\subsubsection{Criteria for Rejecting a Video.}
To ensure the quality of videos, we designed a rejection criteria for annotators to filter the video sources. Each video is simultaneously allocated to at least 5 annotators, and each annotator could independently decide whether to annotate or reject the video. Following~\cite{shou2021generic}, the criteria is designed based on the understandability and the boundary number of the video. Specifically, a video is expected to be rejected in four cases: \textit{(1)} Not understandable due to blurry or overspeeding. \textit{(2)} Contains no boundary or too many boundaries. \textit{(3)} Includes shot changes like zooming, panning or cutting. \textit{(4)} Violating content. The statistics on the number of annotations in all selected videos is shown in Tab.~\ref{table:annotation_num}. We could see that a majority of videos are accepted by at least 5 annotators, indicating the consistency of annotators’ opinions on our annotated videos.

\input{tables/annotation_stat}

\subsubsection{Evaluation of Annotators' Consistency.}
\label{annotation_details:f1_consistency}
Following~\cite{shou2021generic}, we compute F1 score to evaluate the consistency of the annotations towards the same videos. When computing, we take the timestamps of each annotation as the "prediction" and all other annotations in the same video as the "ground truth". Then for each threshold varying from 0.2s to 1s, we compute the precision and recall for the "prediction" to obtain its F1 score. Finally, we average the F1 scores under all thresholds as the final result of the evaluation. The distribution of the average F1 score is shown in Fig.~\ref{fig:consistency_f1}, where over $92\%$ percent of annotations are scored higher than 0.4, suggesting that our annotators have very high consistency in determining event boundary positions. They also tend to focus on the same subject on the agreed boundary, providing captions for the same event change with little bias.

\subsection{Statistics on the Similarity between Captions}

To further investigate the similarity between the captions annotated in the same videos. We first randomly selected 1,000 videos with over 103K captions from our dataset, then computed the Sentence-BERT similarity of status parts’ captions in these videos. The results with three examples for different level of scores are shown in Fig.~\ref{fig:caption_similarity}. We see that nearly 80\% of the caption pairs are less than 0.7 score (only a few words are shared), indicating our captions are unique and fine-grained.

\subsection{Statistics based on the Video Categories in Kinetic-400}

Since we take the “one-step-deeper” events in videos as~\cite{shou2021generic}, the video-level categories in Kinetic-400 could not determine the pattern of events. However, the category provides a higher-level background for our events, thus we conduct further statistics towards it.

\subsubsection{Boundary Number and Interval Duration.}
\begin{figure}[t]
\centering
\includegraphics[width=.99\linewidth]{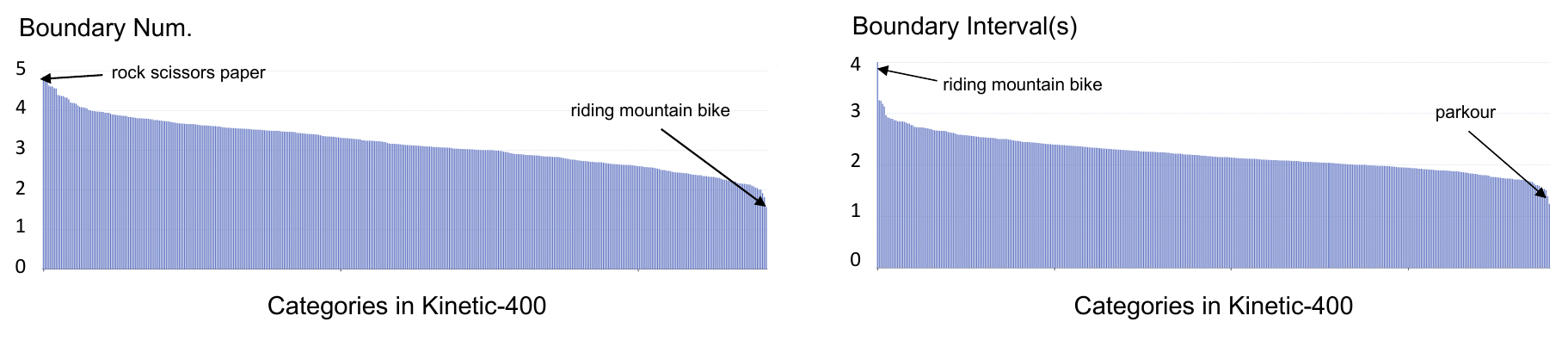}
\caption{\textit{Left.} Average number of boundaries in videos in each \textit{Kinetic-400} category. \textit{Right.} Average duration of boundary intervals in videos in each \textit{Kinetic-400} category.}
\label{fig:boundary_metric_category}
\end{figure}

Firstly we investigate the distribution of boundary numbers in each category of videos. Given a Kinetic-400 category, we compute the average number of boundaries per video in the category. From the result in Fig.~\ref{fig:boundary_metric_category}, we see that the boundary numbers slightly vary with the category and most categories have 2 to 3 boundaries per video. We also illustrate the interval durations versus categories in the right of Fig.~\ref{fig:boundary_metric_category}. In most categories of videos, we could see the average duration of boundary intervals is around 2s.

\subsubsection{Distributions in Splits.}
\begin{figure}[t]
\centering
\includegraphics[width=.99\linewidth]{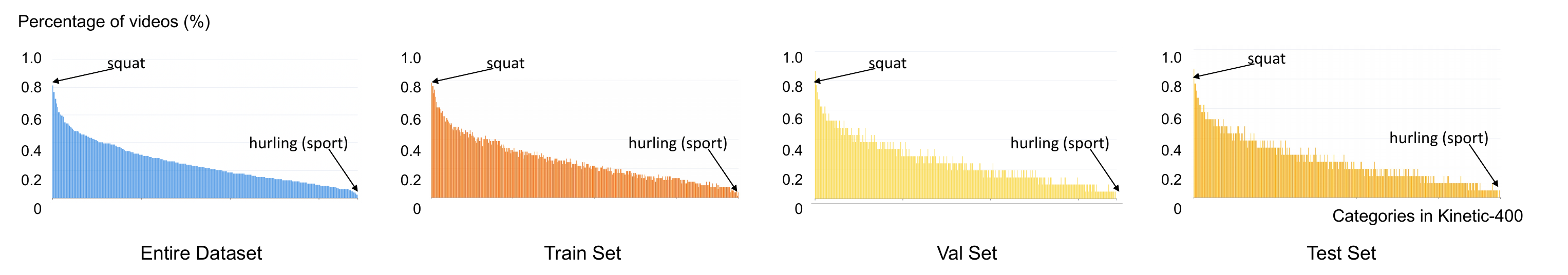}
\caption{Percentage distributions of the videos in each \textit{Kinetic-400} category in the entire dataset and the train/val/test splits. The categories are sorted by their video numbers in the entire dataset}
\label{fig:boundary_category}
\end{figure}

Furthermore, we conduct statistics on the video numbers of each category in our train/val/test splits. The percentage distribution is shown in Fig.~\ref{fig:boundary_category}, where the categories are sorted by their video numbers in the entire dataset. We see that the categories’ distribution in the three splits are consistent with the distribution in the entire dataset.

\subsection{Details of the Adjustment for Downstream Tasks}

In the raw annotation of Kinetic-GEB+, each video is allocated to more than 5 annotators. Due to the variance of annotators’ opinions, the boundary locations in different annotations towards the same video are not the same. When preparing the data for downstream tasks, 
we select one annotator whose labeled boundaries have highest F1 score (computed in Sec.~\ref{annotation_details:f1_consistency}) for each video. 
Then, we use these boundaries' timestamps as the anchors to merge other annotators' captions, preserving the diversity of different opinions. 
Thus, one video corresponds to multiple boundaries, and each boundary could be with multiple captions. 
Finally, we collected 40k anchors/boundaries and from the total 176,681 boundaries in 12,434 videos, where 80\% anchors/boundaries have more than 3 captions and around 10\% of anchors have only 1 unique caption.


As mentioned in main body, the videos in our Kinetic-GEB+ could sometimes contain repeated events or actions, which could disturb the Boundary Grounding task. We found that the difference among some pairs of boundaries within a video is too subtle even for humans to distinguish. Therefore, we need to find these ``equal" pairs of boundaries and mark them as ``equal” boundaries to each other for the Boundary Grounding task. Specifically, when querying with the caption of one boundary in the pair, the timestamps of other ``equal" boundaries are also correct answers. When queried by a boundary caption, the machine is supposed to answer the locations of that boundary as well as all its “equal” boundaries. An example is shown in Fig.~\ref{fig:grounding_res}, where the man changes his status from sitting to standing twice in the video, thus these two status changes are marked as an ``equal" pair.

To find and mark these ``equal" pairs, we employ Sentence-BERT~\cite{reimers2019sentence} to compute the similarity score between the annotated captions of every two different boundaries inside a video. Firstly, we take the filtered annotations. Then for each video, we combine every two of its boundaries to form all the possible pairs. After that, we separate each pair of captions into \textit{subject}, \textit{status before} and \textit{status after} items, and then compute the similarity score for each item using Sentence-BERT. The range of similarity scores is from 0 to 1.
In order to distinguish these ``equal" pairs, we need to set a maximum threshold for similarity scores. First we find that the item pairs scoring less than 0.9 usually have significant differences that are easy for humans to recognize. 
Hence, we collect the pairs of all the three items that score higher than 0.9, and then we annotate manually to classify if each pair is an ``equal" pair. After that, we simulate the decision accuracy of different candidate thresholds varying from 0.9 to 1.0 and finally choose 0.93 as the threshold, where the corresponding accuracy is 95.5\% (i.e. the 2-sigma probability in normal distribution). Finally, we found and marked 4,426 ``equal" pairs consisting of 4,295 boundaries.

\begin{figure}[]
\centering
\includegraphics[width=.99\linewidth]{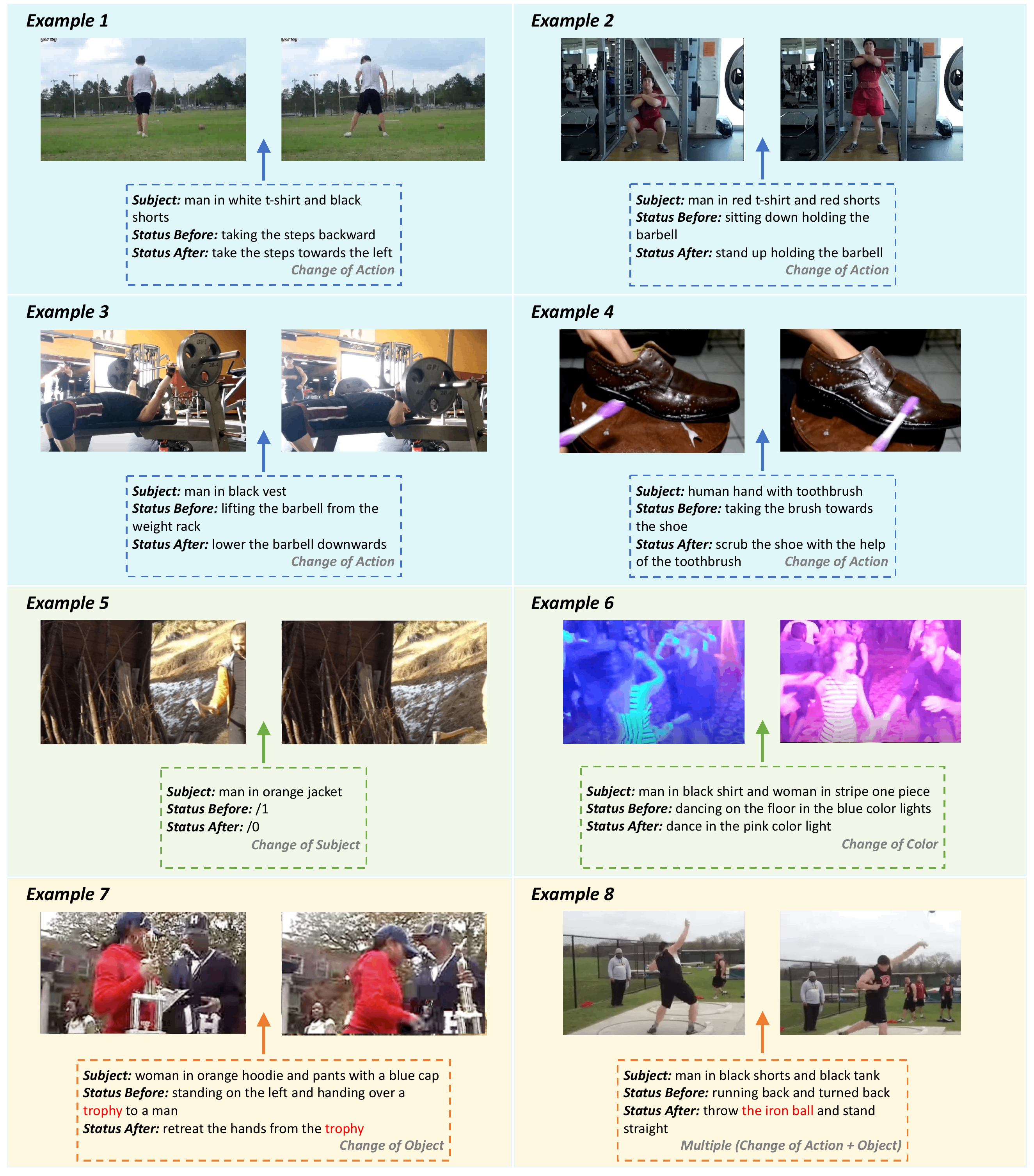}
\caption{More samples from \textit{Kinetic-GEB+} dataset}
\label{fig:supp_dataset_example}
\end{figure}

\subsection{More Examples of Kinetic-GEB+}

Here we illustrate more examples from our Kinetic-GEB+ in Fig.\ref{fig:supp_dataset_example}. The \textit{Example 1} to \textit{4} are all based on Change of Action. The \textit{Example 5} is based on Change of Subject, since the man was at first appearing in the scene and then disappears after the boundary. In \textit{Example 6}, the color of the stage suddenly changes from blue to pink, causing the boundary based on Change of Color. In \textit{Example 7}, the woman was first interacting with the trophy and then retreats her hands to stop interacting after the boundary. This boundary is thus due to the Change of Object being interacted with. Finally in \textit{Example 8}, the boundary is based on Multiple types of status changes. The man in the scene changes his action and simultaneously stops interacting with the iron ball at the boundary.
\section{More Details of Implementation}
\label{supp-sec:imp-detail}

\subsection{Schemes for frame sampling}
In all our experiment groups, if not specified, we employ the two following schemes of frame sampling when extracting visual information for boundary timestamps: 

\textbf{Scheme 1.} In most cases, when using the ground truth boundaries, we set two sampling ranges before and after each boundary timestamp. For the range before the boundary, we set the preceding boundary as the start and the current boundary as the end. Similarly, the range after is between the current boundary and the succeeding boundary. Notably, the predecessor of the first boundary in videos is set to 0, and the successor of the last boundary in videos is set to the end of videos. Finally, we sample 10 frames in each range and 1 frame at the timestamp of the current boundary. This scheme is also employed when using the proposal timestamps generated by GEBD baseline~\cite{shou2021generic}, like in the testing period of Boundary Grounding specified with "GEBD" suffix. 

\textbf{Scheme 2.} Sometimes there is no predefined boundary or proposal, and thus the locations of the preceding and succeeding timestamps are unknown. Therefore, we replace the predecessor and successor with the timestamps 1s before and after the current timestamp. Then we sample 10 frames in each range and 1 frame at the candidate timestamp for further extraction. For example, in the testing period of Boundary Grounding (in the groups without "GEBD" suffix), this scheme is employed by sampling a timestamp candidate every 0.1s for all videos.

\subsection{Further Details in Training}

For each backbone utilized in our experiments, we trained for 50 epochs. For all the BERT based models, we used AdamW optimizer with a linearly decreasing learning rate starting from $5e^{-5}$. Notably, in Boundary Grounding we modify the original contrastive loss in FROZEN~\cite{bain2021frozen} by adding an additional intra loss. Given a batch of embeddings, the intra loss is computed in the same way yet only among the caption and context embeddings from the same videos. Besides, as mentioned in previous sections, we design a batch-random sequential sampler for Boundary Grounding. It ensures more boundaries in the same video to be collected in the same batch, since the boundaries are sequentially sorted by their videos in the dataset. This intra loss and new sampler encourage the model to learn the differences among the boundaries in the same videos, which conforms to the goal of Video Grounding that is selecting the best match among all timestamps in a video.

\subsection{Post-processing and Evaluation}

\begin{figure}[t]
\centering
\includegraphics[width=.99\linewidth]{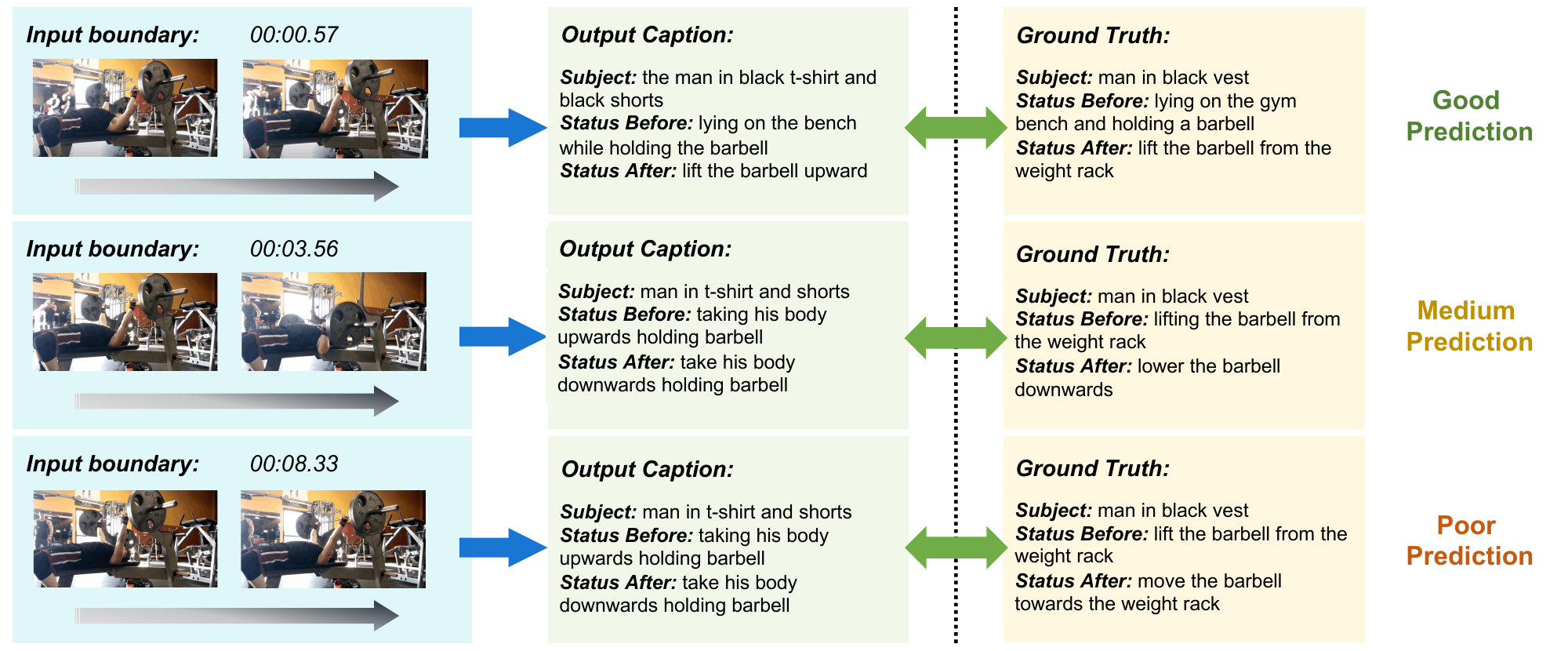}
\setlength{\abovecaptionskip}{0.9cm}
\caption{Samples of Prediction in Boundary Capitioning}
\label{fig:captioning_res}
\end{figure}

\begin{figure}[t]
\centering
\includegraphics[width=.99\linewidth]{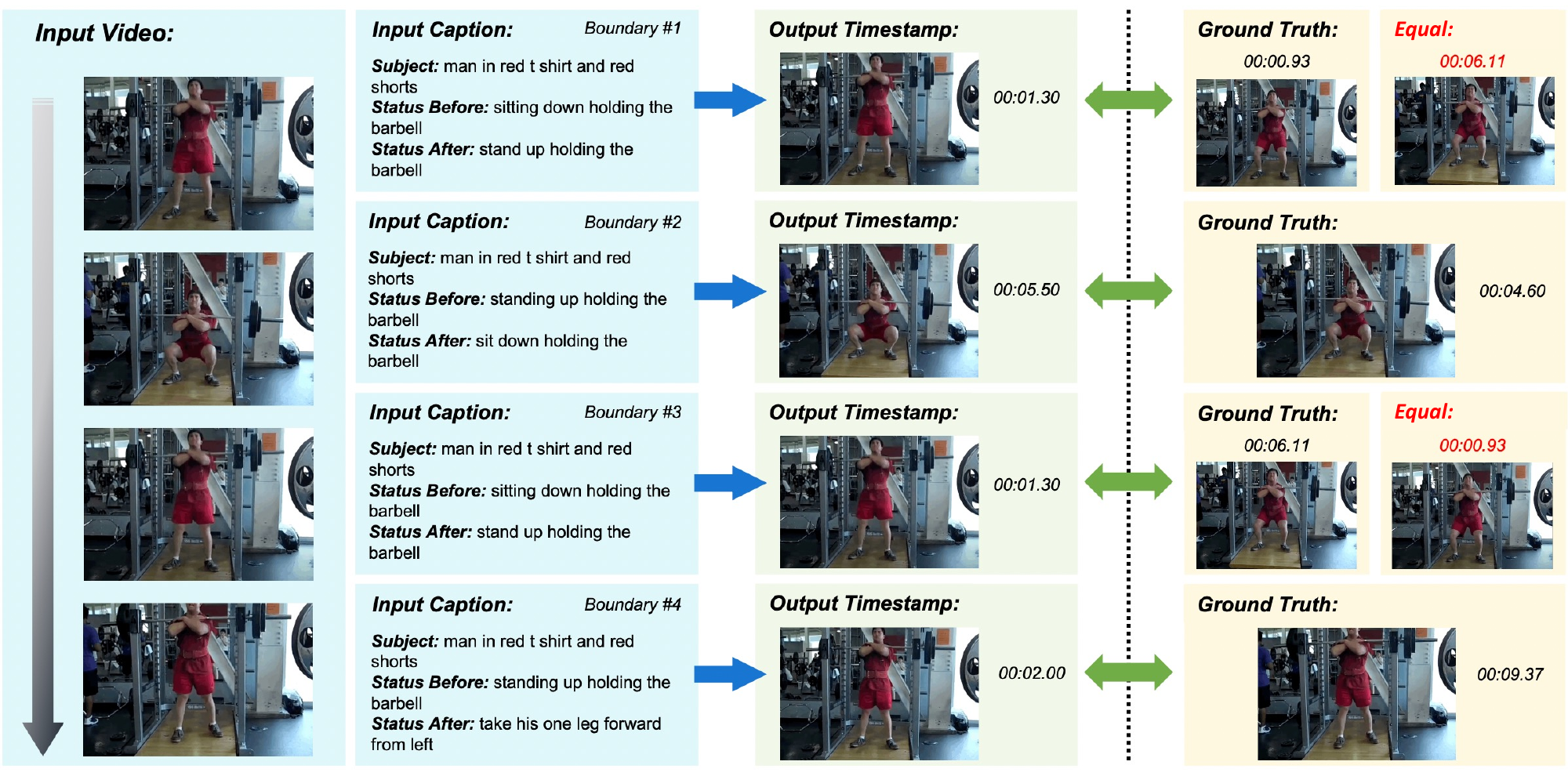}
\setlength{\abovecaptionskip}{0.9cm}
\caption{Samples of Prediction in Boundary Grounding}
\label{fig:grounding_res}
\end{figure}

\begin{figure}[t]
\centering
\includegraphics[width=.97\linewidth]{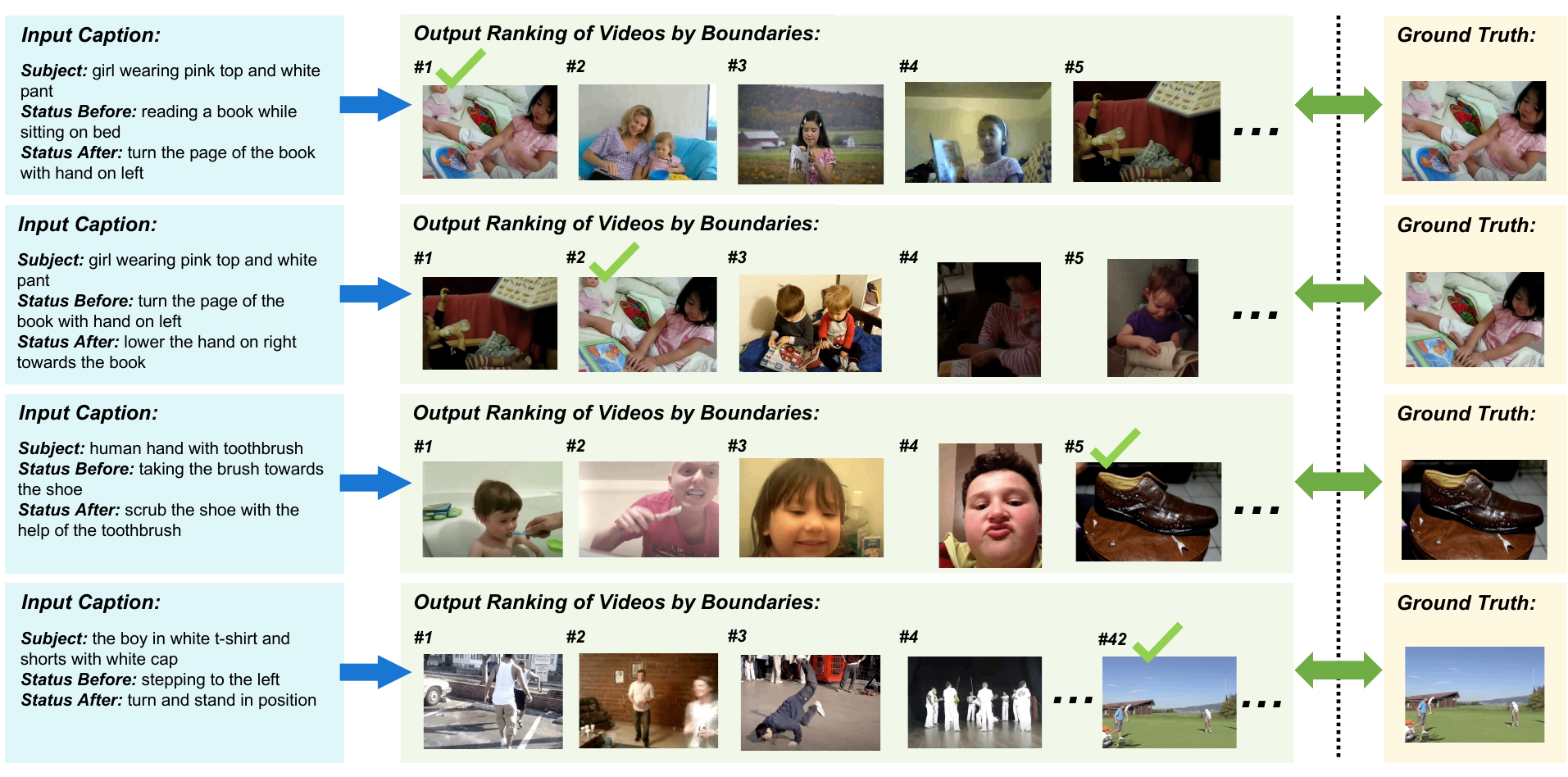}
\vspace{-0.9cm}
\setlength{\abovecaptionskip}{0.9cm}
\caption{Samples of Prediction in Boundary Caption-Video Retrieval}
\label{fig:retrieval_res}
\end{figure}

In Boundary Captioning, we separate and evaluate the \textit{Subject}, \textit{Status Before} and \textit{Status After} items of the generated captions. We found that the conventional BLEU~\cite{papineni-etal-2002-bleu} metric is not suitable for our task and its scores are often inconsistent with humans’ impression, since it only considers the simple repetition of word grams. Samples of predicted captions in a video are illustrated in Fig.~\ref{fig:captioning_res}. We see that the first two generated captions are relatively great, while the caption generated from the last boundary is not satisfying.
For Boundary Grounding, we conduct a post-processing after the models generating the matching scores of all candidates. First we apply the LoG filter~\cite{lindeberg1998feature} to find the local maximas following~\cite{shou2021generic}. Then we select the top-$K$ maximas as final prediction following the statistics of the ground truth timestamp numbers for all queries. After that, we evaluate the finalized prediction by calculating the F1 score under different thresholds, where the computation is the same as in Sec.~\ref{annotation_details:f1_consistency}. Samples of predictions are shown in Fig.~\ref{fig:grounding_res}. Notably, the boundaries at 00:00.93 and 00:06.11 are a pair of equal boundaries, thus we mark both of their timestamps as the ground truths for their caption queries.
For Boundary Caption-Video Retrieval, several samples of predicted ranking are illustrated in Fig.~\ref{fig:retrieval_res}. For the first three samples in the figure, the prediction result is relatively satisfying and the ground truth video is within the top-$5$ of the ranking. However, given the caption of the last sample in the figure, the machine could not clearly recognize the target video from the corpus, and the ground truth video is ranked to \#$42$.


\section{More Exploration on Experiments}
\label{supp-sec:more_experiments}

\subsection{Boundary Captioning}
Here we delve deeper on the design of fusion mechanism for status changes. In Tab.~\ref{table:captioning_more}, we further compare subtraction operation (TPD method) with another 2 operations: 
Simple concatenation (denoted as \textit{Concat}) and \textit{Multimodal Tucker Fusion}~\cite{ben2017mutan}.
As shown, our TPD outperforms the other fusion methods. Furthermore, to investigate the contribution made by different parts in our \textbf{TPD Modeling} method, we conduct an ablation study. In Tab.~\ref{table:captioning_more}, we see that \textit{part a} contributes more than \textit{part b} and \textit{part c}, while the combination of the three parts enables the model to have the best performance.

\begin{table}[t]
\begin{center}
\caption{Results of more exploration on Boundary Captioning, including the comparison among different fusion methods and the ablation study on our \textbf{TPD Modeling} method}
\label{table:captioning_more}
\resizebox{1\linewidth}{!}{
\begin{tabular}{llll|lcccllcccllccc}
\hline
\multicolumn{4}{c|}{\multirow{2}{*}{Method}}      & \multicolumn{4}{c}{CIDEr}                                          &  & \multicolumn{4}{c}{SPICE}                                         &  & \multicolumn{4}{c}{ROUGE\_L}                                      \\
\multicolumn{4}{c|}{}                             & Avg.           & Sub.            & Bef.           & Aft.           &  & Avg.           & Sub.           & Bef.           & Aft.           &  & Avg.           & Sub.           & Bef.           & Aft.           \\ \cline{1-8} \cline{10-13} \cline{15-18} 
\multicolumn{4}{l|}{Concat}   & 68.16          & 86.35           & 62.99          & 55.15          &  & 18.92          & 20.13          & 19.08	          & 17.57          &  & 26.71          & 39.25          & 20.96          & 19.93          \\
\multicolumn{4}{l|}{Tucker}   & 67.25          & 85.15           & 63.08          & 53.51          &  & 18.71          & 20.39          & 18.97          & 16.78          &  & 26.91          & 39.28          & 21.42          & 20.02          \\
\multicolumn{4}{l|}{TPD (part a)}                     & 72.45          & 89.7	           & 70.45          & 57.20          &  & 19.39          & \textbf{20.64}          & 19.87           & 17.67          &  & 27.74          & 39.49          & 22.86          & 20.87          \\
\multicolumn{4}{l|}{TPD (part b)}     & 70.78          & \textbf{90.04}   & 66.59          & 55.70          &  & 19.22          & 20.46	 & 19.67          & 17.54          &  & 27.27          & 39.6	 & 21.93          & 20.27          \\
\multicolumn{4}{l|}{TPD (part c)}   & 69.01          & 86.9           & 66.11          & 54.03          &  & 19.11          & 20.34          & 19.47          & 17.53          &  & 27.31          & \textbf{39.69}          & 21.98          & 20.27          \\ 
\multicolumn{4}{l|}{TPD} & \textbf{74.71} & 85.33           & \textbf{75.98} & \textbf{62.82} &  & \textbf{19.52} & 20.10          & \textbf{20.66} & \textbf{17.81} &  & \textbf{28.15} & 39.16          & \textbf{23.70} & \textbf{21.60} \\ \hline
\end{tabular}
}
\end{center}
\end{table}

\subsection{Boundary Grounding}
In order to investigate the contribution of different captioning parts in Boundary Grounding task, we evaluate some variants of \textit{FROZEN-revised-GEBD} which take no caption (i.e. random guess based on boundary proposals from GEBD) or only subject parts as input for grounding. The F1 scores of \textit{no caption/only subject/full caption} under 0.1s are \textit{3.09/3.25/4.20} respectively. The performance doesn't improve much with only the subject. The reason is that boundaries in the same video are often caused by the same subject, requiring the model to understand captions depicting detailed status changes to ground the video.
\section{More Discussions}
\label{supp-sec:more_discusstion}

\subsection{Common Failures in Boundary Captioning}
\begin{figure}[t]
\centering
\includegraphics[width=.99\linewidth]{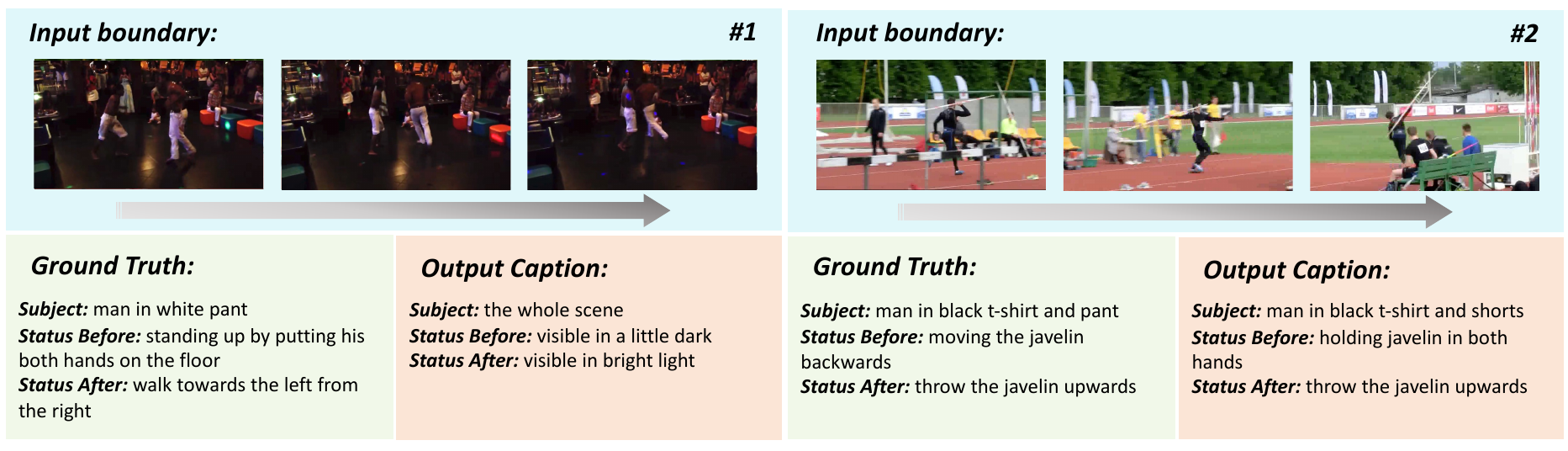}
\setlength{\abovecaptionskip}{0.9cm}
\caption{Two Common Failure Cases in the Task of Boundary Captioning}
\label{fig:failure_case}
\end{figure}

In the task of Boundary Captioning, we find some failure cases happened in our prediction. Here we present two types in Fig.~\ref{fig:failure_case}: (1) the model misses the target subject due to another subject without event change is visually salient. (2) the action in status before or after is subtle and the model mistakenly considers there is nothing happening.

\subsection{``Can we replace Kinetic-GEB+ with existing video captioning datasets simply by concatenating two segments together?''}
It may not work well as (1) previous and next captions from existing datasets could correspond to different subjects while our boundary caption targets one subject; (2) Event caption usually summarises the whole time span, while boundary caption focuses on detailed, fine-grained status change of the subject.

\clearpage

\bibliographystyle{splncs04}
\bibliography{egbib}

\end{document}